\documentclass{article}

\usepackage[final,nonatbib]{neurips_2021}

\usepackage[utf8]{inputenc} 
\usepackage[T1]{fontenc}    
\usepackage{hyperref}       
\usepackage{url}            
\usepackage{booktabs}       
\usepackage{amsfonts}       
\usepackage{nicefrac}       
\usepackage{microtype}      
\usepackage{xcolor}         
\usepackage{cite}

\usepackage{amsmath, amssymb}

\DeclareMathOperator*{\argmin}{arg\,min}
\usepackage{soul}

\usepackage{graphicx}
\graphicspath{ {./figures/} }
\usepackage{algorithm}
\usepackage{algpseudocode}
\usepackage{bbm}

\title{Bubblewrap: Online tiling and real-time flow prediction on neural manifolds}

\author{%
  Anne Draelos \\
  Biostatistics \& Bioinformatics \\
  Duke University \\
  \texttt{anne.draelos@duke.edu} \\
  \And
  Pranjal Gupta \\
  Psychology \& Neuroscience \\
  Duke University \\
  \texttt{pranjal.gupta@duke.edu} \\
  \AND
  Na Young Jun \\
  Neurobiology \\
  Duke University \\
  \texttt{nayoung.jun@duke.edu} \\
  \And
  Chaichontat Sriworarat \\
  Biomedical Engineering \\
  Duke University \\
  \texttt{chaichontat.s@duke.edu} \\
  \And
  John Pearson \\
  Biostatistics \& Bioinformatics \\
  Electrical \& Computer Engineering \\
  Neurobiology \\
  Psychology \& Neuroscience \\
  Duke University \\
  \texttt{john.pearson@duke.edu} \\
}

\begin{document}

\maketitle

\begin{abstract}
  While most classic studies of function in experimental neuroscience have focused on the coding properties of individual neurons, recent developments in recording technologies have resulted in an increasing emphasis on the dynamics of neural populations. This has given rise to a wide variety of models for analyzing population activity in relation to experimental variables, but direct testing of many neural population hypotheses requires intervening in the system based on current neural state, necessitating models capable of inferring neural state online. Existing approaches, primarily based on dynamical systems, require strong parametric assumptions that are easily violated in the noise-dominated regime and do not scale well to the thousands of data channels in modern experiments. To address this problem, we propose a method that combines fast, stable dimensionality reduction with a soft tiling of the resulting neural manifold, allowing dynamics to be approximated as a probability flow between tiles. This method can be fit efficiently using online expectation maximization, scales to tens of thousands of tiles, and outperforms existing methods when dynamics are noise-dominated or feature multi-modal transition probabilities. The resulting model can be trained at kiloHertz data rates, produces accurate approximations of neural dynamics within minutes, and generates predictions on submillisecond time scales. It retains predictive performance throughout many time steps into the future and is fast enough to serve as a component of closed-loop causal experiments.
\end{abstract}

\section{Introduction}
Systems neuroscience is in the midst of a data explosion. Advances in microscopy \cite{ahrens2013whole, emiliani2015all} and probe technology \cite{stevenson2011advances, steinmetz2018challenges, steinmetz2020neuropixels} have made it possible to record thousands of neurons simultaneously in behaving animals. At the same time, growing interest in naturalistic behaviors has increased both the volume and complexity of jointly recorded behavioral data. On the neural side, this has resulted in a host of new modeling and analysis approaches that aim to match the complexity of these data, typically using artificial neural network models as proxies for biological neural computation \cite{mante2013context, rajan2016recurrent, song2016training}. 

At the same time, this increase in data volume has resulted in increasing emphasis on methods for dimensionality reduction \cite{cunningham2014dimensionality} and a focus on neural populations in preference to the coding properties of individual neurons \cite{ebitz2021population}. However, given the complexity of neural dynamics, it remains difficult to anticipate what experimental conditions will be needed to test population hypotheses in \emph{post hoc} analyses, complicating experimental design and reducing power. Conversely, adaptive experiments, those in which the conditions tested change in response to incoming data, have been used in neuroscience to optimize stimuli for experimental testing \cite{carlson2011sparse, dimattina2013adaptive, NIPS2017_892c91e0, abbasi2018deeptune}, in closed-loop designs \cite{zhang2018closed, bolus2020state,peixoto2021decoding}, and even to scale up holographic photostimulation for inferring functional connectivity in large circuits \cite{draelos2020online}. 

Yet, despite their promise, adaptive methods are rarely applied in practice for two reasons: First, although efficient online methods for dimensionality reduction exist \cite{brand2002incremental,baker2004block,brand2006fast,baker2012low,mairal2010online}, these methods do not typically identify \emph{stable} dimensions to allow low-dimensional representations of data to be compared across time points. That is, when the spectral properties of the data are changing in time, methods like incremental SVD may be projecting the data into an unstable basis, rendering these projections unsuitable as inputs to further modeling. Second, while many predictive models based on the dynamical systems approach exist \cite{mante2013context, archer2015black,gao2016linear,pandarinath2018inferring, linderman2017bayesian,linderman2017recurrent,nassar2018tree}, including online approaches \cite{NIPS2016_b2531e7b,yang2018control,bolus2020state,zhao2020variational}, they typically assume a system with lawful dynamics perturbed by Gaussian noise. However, many neural systems of interest are noise-dominated, with multimodal transition kernels between states. 

In this work, we are specifically interested in closed loop experiments in which predictions of future neural state are needed in order to time and trigger interventions like optogenetic stimulation or a change in visual stimulus. Thus, our focus is on predictive accuracy, preferably far enough into the future to compensate for feedback latencies. To address these goals, we propose an alternative to the linear systems approach that combines a fast, stable, online dimensionality reduction with a semiparametric tiling of the low-dimensional neural manifold. This tiling introduces a discretization of the neural state space that allows dynamics to be modeled as a Hidden Markov Model defined by a sparse transition graph. The entire model, which we call ``Bubblewrap,'' can be learned online using a simple EM algorithm and handles tilings and graphs of up to thousands of nodes at kiloHertz data rates. Most importantly, this model outperforms methods based on dynamical systems in high-noise regimes when the dynamics are more diffusion-like. Training can be performed at a low, fixed latency $\approx$10ms using a GPU, while a cached copy of the model in main memory is capable of predicting upcoming states at $<$1ms latency. As a result, Bubblewrap offers a method performant and flexible enough to serve as a neural prediction engine for causal feedback experiments.

\section{Stable subspaces from streaming SVD}
As detailed above, one of the most pressing issues in online neural modeling is dealing with the increasingly large dimensionality of collected data --- hundreds of channels per Neuropixels probe \cite{steinmetz2018challenges, steinmetz2020neuropixels}, tens of thousands of pixels for calcium imaging. However, as theoretical work has shown \cite{gao2017theory,trautmann2019accurate}, true neural dynamics often lie on a low-dimensional manifold, so that population activity can be accurately captured by analyzing only a few variables. 

Here, we combine two approaches to data reduction: In the first stage, we use sparse random projections to reduce dimensionality from an initial $d$ dimensions (thousands) to $n$ (a few hundred) \cite{achlioptas2003database,li2006very}. By simple scaling, for a fixed budget of $N$ cells in our manifold tiling, we expect density (and thus predictive accuracy) to scale as $N^{\frac{1}{n}}$ in dimension $n$, and so we desire $n$ to be as small as possible. However, by the Johnson-Lindenstrauss Lemma \cite{johnson1984extensions, li2006very}, when reducing from $d$ to $n$ dimensions, the distance between vectors $u_*$ and $v_*$ in the reduced space is related to the distance between their original versions $u$ and $v$ by 
\begin{equation}
  \label{jl_lemma}
	(1 - \varepsilon)\lVert u - v \rVert^2 \le \lVert u_* - v_* \rVert^2 \le (1 + \varepsilon)\lVert u - v \rVert^2
\end{equation}
with probability $1 - \delta$ if $n > \mathcal{O}(\log (1/\delta)/\varepsilon^2)$. Unfortunately, even for $\varepsilon \sim 0.1$ (10\% relative error), the required $n$ may be quite large, making this inappropriate for reducing to the very small numbers of effective dimensions characterizing neural datasets. 

\begin{figure}
  \centering
  \includegraphics[width=\linewidth]{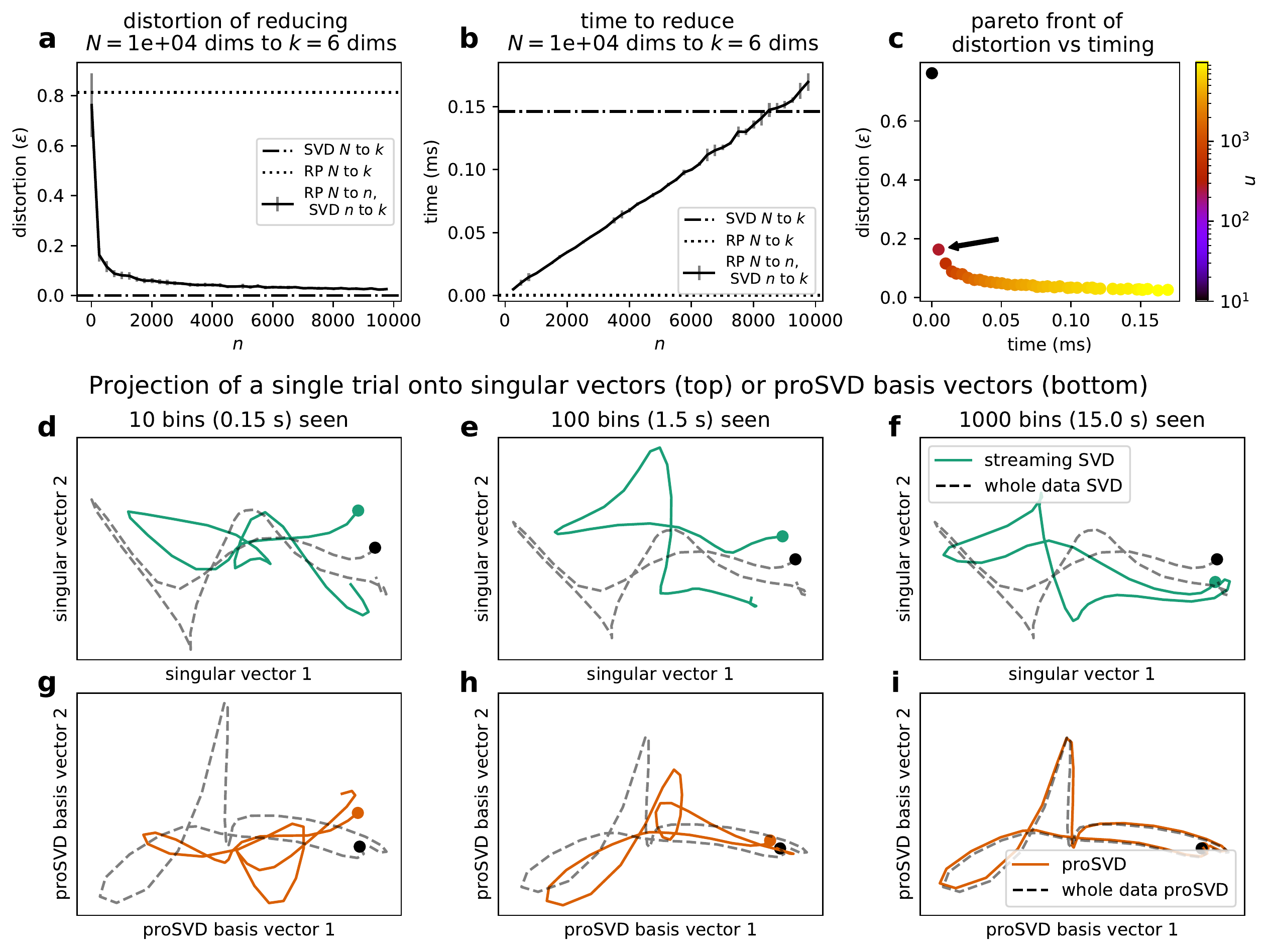}
  \caption{\textbf{Timing and stability of two-stage dimension reduction.} \textbf{a)} Distortion ($\varepsilon$) as a function of number of dimensions retained ($n$) for both sparse random projections and proSVD on random Gaussian data with batch size $b=1000$. \textbf{b)} Time required for the dimensionality reduction in \textbf{(a)}, amortized for batch size. While random projections are extremely efficient, proSVD time costs grow with the number of dimensions retained. \textbf{c)} Pareto front for the time-distortion tradeoff of random projections followed by proSVD. Color indicates $n$, the number of dimensions retained by random projections. Black arrow indicates the particular tradeoff we chose of $n = 200$. \textbf{d--f)} Embedding of a single trial (green line) into the basis defined by streaming SVD for different amounts of data seen. Dotted line indicates the same trial embedded using SVD on the full data set. Rapid changes in estimates of singular vectors early on lead to an unstable representation. \textbf{g--i)} Same trial and conventions as \textbf{(d--f)} for the proSVD embedding. Dotted lines in the two rows are the same curve in different projections.} 
  \label{fig:prosvd}
\end{figure}

Thus, in the second stage, we reduce from $n\sim \mathcal{O}(100)$ to $k \sim \mathcal{O}(10)$ dimensions using a streaming singular value decomposition. This method is based on the incremental block update method of \cite{baker2004block,baker2012low} with an important difference: While the block update method aims to return the top-$k$ SVD at every time point, the directions of the singular vectors may be quite variable during the course of an experiment (Figure \ref{fig:prosvd}d--h), which implies an unstable representation of the neural manifold. However, as we show below, the top-$k$ \emph{subspace} spanned by these vectors stabilizes in seconds on typical neural datasets and remains so throughout the experiment. Therefore, by selecting a stable basis (instead of the singular vector basis) for the top-$k$ subspace, we preserve the same information while ensuring a stable representation of the data for subsequent model fitting.

More specifically, let $\mathbf{x}_t \in \mathbb{R}^n$ be a vector of input data after random projections. In our streaming setup, these are processed $b$ samples at a time, with $b=1$ reasonable for slower methods like calcium imaging and $b=40$ more appropriate for electrophysiological sampling rates of $\sim$20kHz. Then, if the data matrix $X$ has dimension $n \times T$, adding columns over time, the incremental method of \cite{baker2004block,baker2012low} produces at each time step a factorization $X = QRW^\top$, where the columns of the orthogonal matrices $Q$ and $W$ span the left and right top-$k$ singular subspaces, respectively. If the matrix $R$ were diagonal, this would be equivalent to the SVD. In the incremental algorithm, $R$ is augmented at each timestep based on new data to form $\hat{R}$, which is block diagonalized via an orthogonal matrix and truncated to the top-$k$ subspace, allowing for an exact reduced-rank SVD (Appendix \ref{app:prosvd-details}). 

However, as reviewed in \cite{baker2004block,baker2012low}, since there are multiple choices of basis $Q$ for for the top-$k$ singular subspace, there are likewise multiple choices of block diagonalization for $\hat{R}$. In \cite{baker2004block,baker2012low}, the form of this operation is chosen for computational efficiency. But an equally valid option is to select the orthogonal matrix that minimizes the change in the singular subspace basis $Q$ from one timestep to the next:
\begin{equation}
  \min \lVert Q_{t} - Q_{t-1}\rVert_F = \min_T \lVert \hat{Q} U_1 T^\top - Q_{t-1} \rVert_F ,
\end{equation} 
where $\hat{Q}$ is an augmented basis for the top-$(k+b)$ singular subspace, $U_1$ contains the first $k$ left singular vectors of $\hat{R}$, and $T$ is an orthogonal matrix (Appendix \ref{app:prosvd-details}). This minimization is known as the Orthogonal Procrustes problem and has a well-known solution \cite{schonemann1966generalized}: $T = \tilde{U}\tilde{V}^\top$, where $\tilde{U}$ and $\tilde{V}$ are the left and right singular vectors, respectively, of $M \equiv Q^\top_{t-1}\hat{Q}U_1$. (See \cite{degenhart2020stabilization} for a recent application of similar ideas in brain-computer interfaces). This Procrustean SVD (proSVD) procedure is summarized in Algorithm \ref{alg:prosvd}. There, lines 1--8 follow \cite{baker2004block,baker2012low}, while lines 10 and 11 perform the Orthogonal Procrustes procedure. Line 9 serves as a leak term that discounts past data as in \cite{ross2008incremental}.

\begin{algorithm}[t]
  \caption{Procrustean SVD (proSVD)}
  \label{alg:prosvd}
  \begin{algorithmic}[1]
  \State {\bf Given:} Initial data $X_0$, decay parameter $\alpha \in (0, 1]$ 
  \State {\bf Initialize:} QR Factorization: $X_0 = Q_0 R_0$ 
  \State
  \For{$t=1\ldots$}
  \State Fetch $b$ new columns of data, $X_+$ 
  \State $C \gets Q_{t-1}^\top X_+$, \quad $X_\perp \gets X_+ - Q_{t-1}C$, \quad $Q_\perp, R_\perp \gets \mathtt{QR}(X_\perp)$ \Comment{Gram-Schmidt}
  \State $\hat{Q} \gets \begin{bmatrix} Q_{t-1} & Q_\perp \end{bmatrix}$, \quad $\hat{R} \gets \begin{bmatrix} R_{t-1} & C \\ 0 & R_\perp \end{bmatrix}$ \Comment{QR of augmented data}
  \State $U, \Sigma, V \gets \mathtt{SVD}(\hat{R})$ 
  \State $\Sigma \gets \alpha\Sigma$ \Comment{Discount old data}
  \State $M \gets Q^\top_{t-1}\hat{Q}U_1 = \begin{bmatrix} \mathbbm{1}_{k\times k} & \mathbf{0}_{k \times b} \end{bmatrix}U_1$ \Comment{$U_1$ contains the first $k$ columns of $U$}
  \State $\tilde{U}, \tilde{\Sigma}, \tilde{V} \gets \mathtt{SVD}(M)$, \quad $T \gets \tilde{U}\tilde{V}^\top$ \Comment{Orthogonal Procrustes: $\min_T \lVert \hat{Q}U_1T^\top - Q_{t-1}\rVert_F$}
  \State $Q_t \gets \hat{Q}U_1 T^\top$ \Comment{Update left subspace basis}
  \State $Q_v, R_v \gets \mathtt{QR}(V)$, \quad $R_t \gets T\Sigma_1 Q_v^\top$ \Comment{QR right subspace, update inner block}
  \EndFor
  \end{algorithmic}
\end{algorithm}

Figures \ref{fig:prosvd}a-c illustrates the performance of the two-stage dimension reduction algorithm for a case of $d=10^4$ randomly generated Gaussian data. While proSVD yields minimal distortion (due to truncation of the spectrum to $k=6$), random projections require $k \sim \mathcal{O}(100)$ to achieve the same result (Figure \ref{fig:prosvd}a). By contrast, random projections are much faster (Figure \ref{fig:prosvd}b). Thus, we can trade off distortion against time by adjusting $n$, the number of intermediate dimensions. As Figure \ref{fig:prosvd}c shows, the optimal tradeoff occurs somewhere around $n=200$ for this example.

Figures \ref{fig:prosvd}d-i show results for neural data from recorded from monkey motor cortex \cite{pandarinath2018inferring} in a cued reach task. While projection of the data into the basis defined by streaming SVD remains unstable early in data collection (top), the proSVD representation is nearly equivalent to the full offline result after only a few trials ($\approx$15s, middle). This is due to the fact that, in all data sets we examined, the top-$k$ SVD \emph{subspace} was identified extremely quickly; proSVD simply ensures the choice of a stable basis for that subspace. 

\section{Bubblewrap: a soft manifold tiling for online modeling}
As reviewed above, most neural population modeling approaches are based on the dynamical systems framework, assuming a lawful equation of motion corrupted by noise. However, for animals engaged in task-free natural behavior \cite{berman2014mapping,berman2016predictability, pereira2020quantifying}, trajectories are likely to be sufficiently complex that simple dynamical models fail. For instance, dynamical systems models with Gaussian noise necessarily produce unimodal transition probabilities centered around the mean prediction, while neural trajectories may exhibit multimodal distributions beginning at the same system state. By contrast, we pursue an alternative method that trades some accuracy in estimating instantaneous system state for flexibility in modeling the manifold describing neural activity. 

Our approach is to produce a soft tiling of the neural manifold in the form of a Gaussian mixture model (GMM), each component of which corresponds to a single tile. We then approximate the transitions between tiles via a Hidden Markov Model (HMM), which allows us to capture multimodal probability flows. As the number of tiles increases, the model produces an increasingly finer-grained description of dynamics that assumes neither an underlying dynamical system nor a particular distribution of noise.

More specifically, let $x_t$ be the low-dimensional system state and let $z_t \in 1\ldots N$ index the tile to which the system is assigned at time $t$. Then we have for the dynamics
\begin{equation}
  \label{GMM_HMM}
  p(z_{t}=j| z_{t-1}=i) = A_{ij} \quad p(x_{t}|z_{t}) = \mathcal{N}(\mu_{z_t}, \Sigma_{z_t}) \quad p(\mu_j, \Sigma_j) = \mathrm{NIW}(\mu_{0j}, \lambda_j, \Psi_j, \nu_j) ,
\end{equation}
where we have assumed Normal-inverse-Wishart priors on the parameters of the Gaussians. Given its exponential family form and the conjugacy of the priors, online expectation maximization updates are available in closed form \cite{mongillo2008online,cappe2009line,le2013online} for each new datum, though we opt, as in \cite{cappe2009line} for a gradient-based optimization of an estimate of the evidence lower bound
\begin{align}
  \label{Q_GMM_HMM}
  \mathcal{L}(A, \mu, \Sigma) &= \sum_{ij} (\hat{N}_{ij}(T) + \beta_{ij} - 1)\log A_{ij} + \sum_j (\hat{S}_{1j}(T) + \lambda_j \mu_{0j})^\top \Sigma_j^{-1}\mu_j \\
  &\phantom{=} - \frac{1}{2}\sum_j \mathrm{tr}((\Psi_j + \hat{S}_{2j}(T) + \lambda_j \mu_{0j} \mu_{0j}^\top + (\lambda_j + \hat{n}_j(T))\mu_j \mu_j^\top)\Sigma_j^{-1}) \nonumber \\
  &\phantom{=} -\frac{1}{2}\sum_j (\nu_j + \hat{n}_j(T) + d + 2) \log \det \Sigma_j  \nonumber 
\end{align}
with accumulating (estimated) sufficient statistics
\begin{align}
  \label{S_update}
  \alpha_j(t) &= \sum_i \alpha_i(t-1)\Gamma_{ij}(t) & 
  \hat{N}_{ij}(t) &= (1 - \varepsilon_{t})\hat{N}_{ij}(t-1) + \alpha_i(t-1)\Gamma_{ij}(t) \\ 
  \hat{n}_j(t) &= \sum_i \hat{N}_{ij}(t) &
  \hat{S}_{1j}(t) &= (1 - \varepsilon_{t})\hat{S}_{1j}(t-1) + \alpha_j(t) x_t \nonumber \\
  & &
  \hat{S}_{2j}(t) &= (1 - \varepsilon_{t})\hat{S}_{2j}(t-1) + \alpha_j(t) x_t x_t^\top \nonumber
\end{align}
where $\alpha_j(t) = p(z_t=j|x_{1:t})$ is the filtered posterior, $\Gamma_{ij}(t)$ is the update matrix from the forward algorithm \cite{mongillo2008online}, and $\varepsilon_t$ is a forgetting term that discounts previous data. Note that even for $\varepsilon = 0$, $\mathcal{L}$ is only an estimate of the evidence lower bound because the sufficient statistics are calculated using $\alpha(t)$ and not the posterior over all observed data.

In setting Normal-Inverse-Wishart priors over the Gaussian mixture components, we take an empirical Bayes approach by setting prior means $\mu_{0j}$ to the current estimate of the data center of mass and prior covariance parameters $\Psi_j$ to $N^{-\frac{2}{k}}$ times the current estimate of the data covariance (Appendix \ref{app:bubblewrap_details}). For initializing the model we use a small data buffer $M \sim \mathcal{O}$(10). We chose effective observation numbers $(\lambda, \nu) = 10^{-3}$ and trained this model to maximize $\mathcal{L}(A, \mu, \Sigma)$ using Adam \cite{kingma2014adam}, enforcing parameter constraints by replacing them with unconstrained variables $a_{ij}$ and lower triangular $L_j$ with positive diagonal: $A_{ij} = \exp(a_{ij})/\sum_j \exp(a_{ij})$, $\Sigma_j^{-1} = L_j L_j^\top$. 

Finally, in order to prevent the model from becoming stuck in local minima and to encourage more effective tilings, we implemented two additional heuristics as part of Bubblewrap: First, whenever a new observation was highly unlikely to be in any existing mixture component ($\log p(x_t|z_t) < \theta_n$ for all $z_t$), we teleported a node at this data point by assigning $\alpha_J(t) = 1$ for an unused index $J$. For initial learning this results in a ``breadcrumbing'' approach where nodes are placed at the locations of each new observed datum. Second, when the number of active nodes was equal to our total node budget $N$, we chose to reclaim the node with the lowest value of $\hat{n}(t)$ and zeroed out its existing sufficient statistics before teleporting it to a new location. In practice, these heuristics substantially improved performance, especially early in training (Appendix \ref{app:additional-exp}). The full algorithm is summarized in Algorithm \ref{alg:bubblewrap}.

\begin{algorithm}[h]
  \caption{Bubblewrap}
  \label{alg:bubblewrap}
  \begin{algorithmic}[1]
  \State {\bf Given:} Hyperparameters $\lambda_j, \nu_j, \beta_{t}$, forgetting rate $\varepsilon_t$, teleport threshold $\theta$, step size $\delta$, initial data buffer $M$
  \State {\bf Initialize} with $\lbrace x_1\ldots x_M\rbrace${\bf :} $\mu_j \gets \bar{\mu}$, $\Sigma_j \gets \bar{\Sigma}$, $a_{ij} \gets \frac{1}{N}$, $\alpha_j \gets \lambda_j \frac{1}{N}$.
  \State
  \For{$t=1\ldots$}
  \State Observe new data point $x_t$.
  \If{$\log p(x_t|z_t) < \theta$ $ \forall z_t$} \Comment{Teleport}
  \State $\mu_J = x_t, \alpha_J(t) = 1$ for $J = \argmin_j \hat{n}_j(t)$
  \EndIf
  \State Calculate $\Gamma_{ij}(t)$ via forward filtering \cite{mongillo2008online}.
  \State Update sufficient statistics via (\ref{S_update}). \Comment{E step}
  \State $\bar{\mu} \gets \frac{\sum_j \hat{S}_{1j}}{\sum_j \hat{n}_j}$, $\overline{\Sigma} \gets \frac{\sum_j \hat{S}_{2j}}{\sum_j \hat{n}_j} - \bar{\mu}\bar{\mu}^T$ \Comment{Global mean and covariance update}
  \State $\epsilon_j \sim \mathcal{N}(0, \eta^2)$, $\mu_{0j} \gets a \mu_{0j} + (1 - a)\bar{\mu} + \epsilon_j$, $\Psi_j \gets \frac{\overline{\Sigma}}{N^{\frac{2}{k}}}$ \Comment{Update priors (Appendix \ref{app:bubblewrap_details})}
  \State Perform gradient-based update of $\mathcal{L}(A, \mu, \Sigma)$ (\ref{Q_GMM_HMM}) \Comment{M step}
  \EndFor
  \end{algorithmic}
\end{algorithm}

\section{Experiments}

We demonstrated the performance of Bubblewrap on both simulated non-linear dynamical systems and experimental neural data. We compared these results to two existing online learning models for neural data, both of which are based on dynamical systems \cite{NIPS2016_b2531e7b,zhao2020variational}. To simulate low-dimensional systems, we generated noisy trajectories from a two-dimensional Van der Pol oscillator and a three-dimensional Lorenz attractor. For experimental data, we used four publicly available datasets from a range of applications: 1) trial-based spiking data recorded from primary motor cortex in monkeys performing a reach task \cite{churchlandwebsite, churchland2012neural} preprocessed by performing online jPCA \cite{churchland2012neural}; 2) continuous video data and 3) trial-based wide-field calcium imaging from a rodent decision-making task \cite{musall2019single,musalldata}; 4) high-throughput Neuropixels data \cite{steinmetz_pachitariu_stringer_carandini_harris_2019, stringer2019spontaneous}. 

For each data set, we gave each model the same data as reduced by random projections and proSVD. For comparisons across models, we quantified overall model performance by taking the mean log predictive probability over the last half of each data set (Table \ref{res-table}). For Bubblewrap, prediction $T$ steps into the future gives
\begin{align}
  \log p(x_{t + T}|x_{1:t}) &= \log \sum_{i,j} p(x_{t+T}|z_{t+T}=j) p(z_{t+T}=j|z_t=i) p(z_t=i|x_{1:t}) \nonumber \\
  &= \log \sum_{i,j} \mathcal{N}(x_{t+1}; \mu_j, \Sigma_j) (A^T)_{ij} \alpha_i(t) \label{eqn:bubblewrap_pred},
\end{align}
where $A^T$ is the $T$-th power of the transition matrix.
Conveniently, these forward predictions can be efficiently computed due to the closed form (\ref{eqn:bubblewrap_pred}), while similar predictions in comparison models \cite{NIPS2016_b2531e7b, zhao2020variational} must be approximated by sampling (Appendix \ref{app:predictive_calcs}).
In addition, for Bubblewrap, which is focused on coarser transitions between tiles, we also report the entropy of predicted transitions:
\begin{align}
  H(t,T) &= -\sum_j p(z_{t+T}=j|x_{1:t}) \log p(z_{t+T}=j|x_{1:t}) = - \sum_{ij} (A^T)_{ij}\alpha_i(t) \log \sum_k (A^T)_{kj}\alpha_k(t) .
\end{align}
Additional detailed experimental results and benchmarking of our GPU implementation in JAX \cite{jax2018github} are in Appendix \ref{app:additional-exp}. We compared performance of our algorithm against both \cite{NIPS2016_b2531e7b} (using our own implementation in JAX) and Variational Joint Filtering \cite{zhao2020variational} (using the authors' implementation). Our implementation of Bubblewrap, as well as code to reproduce our experiments, is open-source and available online at
\url{http://github.com/pearsonlab/Bubblewrap}.

\begin{table}
  \centering
  \caption{Model comparison results as mean $\pm$ standard deviation of the log predictive probability over the last half of the dataset. Asterisks (*) indicate models that degenerated to a random walk.}
  \begin{tabular}{llll}
    \toprule
     & \multicolumn{3}{c}{Log predictive probability}                   \\
    \cmidrule(r){2-4}
    Dataset & Bubblewrap & VJF \cite{zhao2020variational} & ZP (2016) \cite{NIPS2016_b2531e7b} \\
    \midrule
    2D Van der Pol, 0.05 & $\phantom{-}0.965\pm1.123$ & $-0.338\pm0.427$ & $\phantom{-}0.121\pm0.857$\\
    2D Van der Pol, 0.20 & $ -1.088\pm1.184$ & $-1.140\pm0.879$ & $-0.506\pm0.964$\\
    3D Lorenz, 0.05 &  $-7.338\pm1.289$ & $-16.98\pm1.923$  & $-12.39\pm1.723 * $   \\
    3D Lorenz, 0.20 & $-7.474 \pm 1.279$ & $-17.30\pm2.112$ & $-12.42\pm1.708 * $   \\
    Monkey reach & $\phantom{-}3.046\pm4.959$ & $-5.159\pm0.987$ & $\phantom{-}3.818\pm9.118$  \\
    Wide-field calcium & $\phantom{-}5.974\pm2.979$ & $\phantom{-}3.768\pm6.204$ & $\phantom{-}1.613\pm4.083$  \\
    Mouse video &  $-10.93\pm2.386$ & $-15.86\pm1.084$ & $-10.65\pm4.145 *$  \\
    Neuropixels & $-12.84\pm6.017$ & $-12.06\pm5.244$ & $-12.28\pm4.567$  \\
    \bottomrule
  \end{tabular}
  \label{res-table}
\end{table}

\begin{figure}
  \centering
  \includegraphics[width=\linewidth]{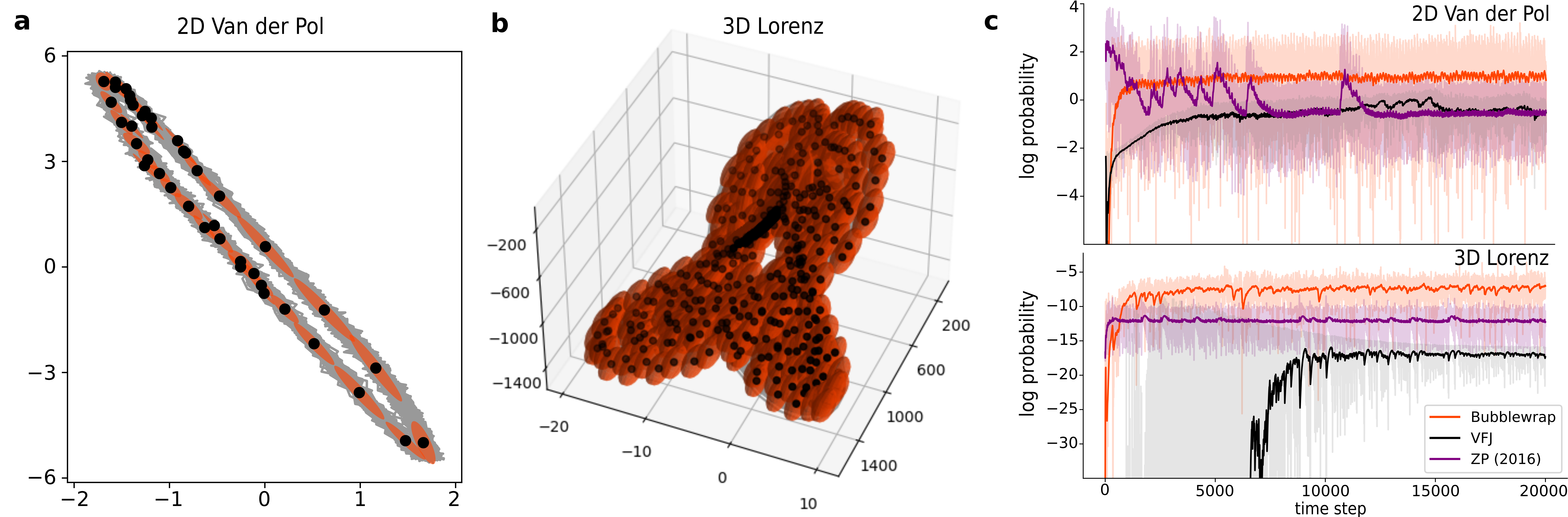}
  \caption{\textbf{Modeling of low-dimensional dynamical systems.} \textbf{a)} Bubblewrap end tiling of a 2D Van der Pol oscillator (data in gray; 5\% noise case corresponding to line 1 of Table \ref{res-table}). Tile center locations are in black with covariance 'bubbles' for 3 sigma in orange.  \textbf{b)} Bubblewrap end tiling of a 3D Lorenz attractor (5\% noise), where tiles are plotted similarly to (a). \textbf{c)} Log predictive probability across all timepoints for each comparative model for the 2D Van der Pol, 0.05 case (top) and for the 3D Lorenz, 0.05 case (bottom). } 
  \label{fig:2d3d}
\end{figure}

When tested on low-dimensional dynamical systems, Bubblewrap successfully learned tilings of both neural manifolds, outperforming VJF \cite{zhao2020variational} on both datasets (Figure \ref{fig:2d3d}a,b) while it was comparable to the algorithm of \cite{NIPS2016_b2531e7b} on one of the 2D (but neither of the 3D) cases (Figure \ref{fig:2d3d}). This is surprising, since both comparison methods assume an underlying dynamical system and attempt to predict differences between individual data points, while Bubblewrap only attempts to localize data to within a coarse area of the manifold.

\begin{figure}
  \centering
  \includegraphics[width=\linewidth]{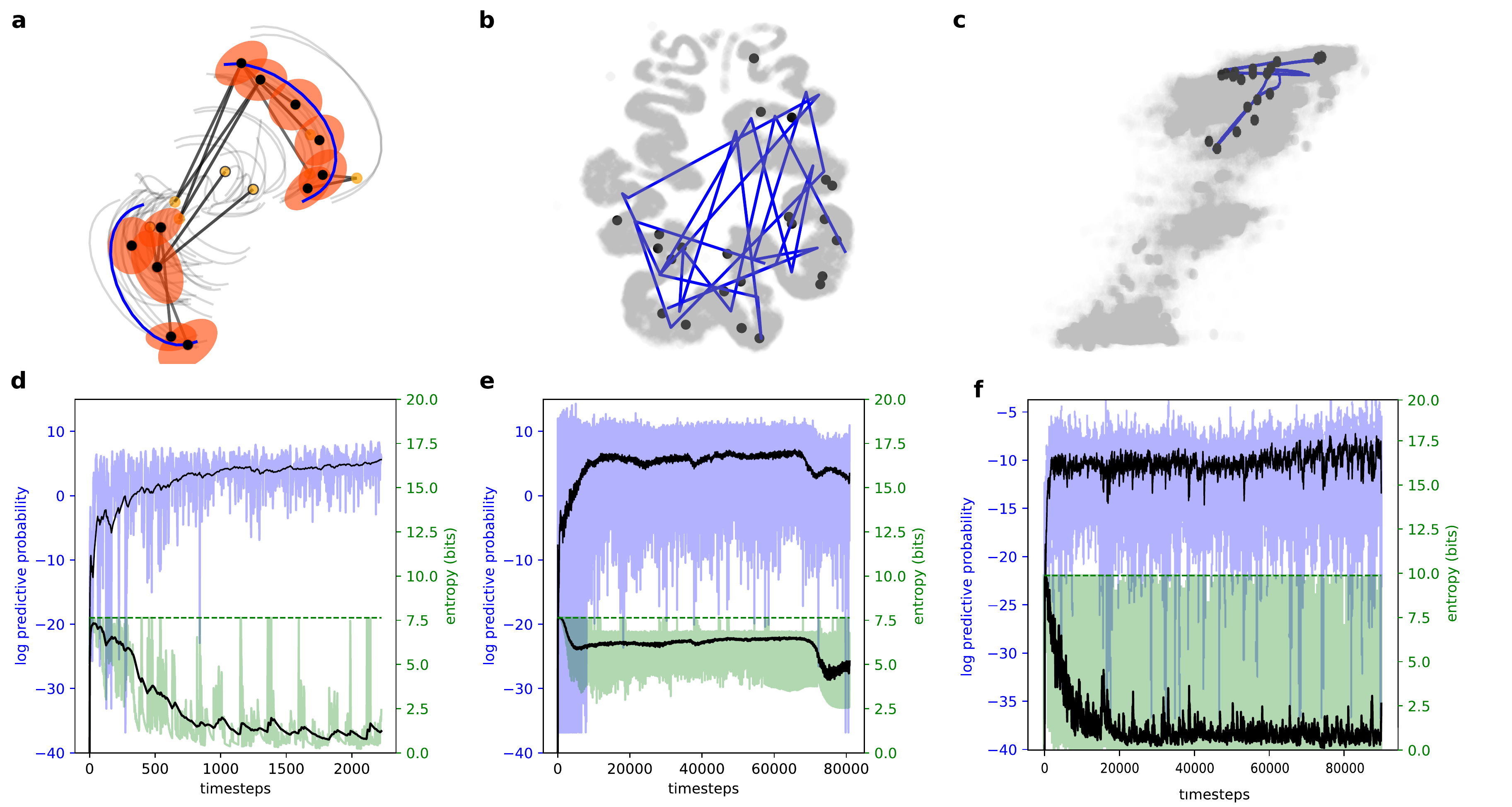}
  \caption{\textbf{Bubblewrap results on experimental datasets.} \textbf{a)} Bubblewrap results for example trials (blue) from the monkey reach dataset \cite{churchlandwebsite, churchland2012neural}, projected onto the first jPCA plane. All trials are shown in gray. The tile center locations which were closest to the trajectories are plotted along with their covariance "bubbles." Additionally, large transition probabilities from each tile center are plotted as black lines connecting the nodes.  Bubblewrap learns both within-trial and across-trial transitions, as shown by the probability weights.  \textbf{b)} Bubblewrap results on widefield calcium imaging from \cite{musall2019single, musalldata}, visualized with UMAP. A single trajectory comprising $\approx$1.5s of data is shown in blue. Covariance "bubbles" and transition probabilities omitted for clarity. \textbf{c)} Bubblewrap results when applied to videos of mouse behavior \cite{musall2019single, musalldata}, visualized by projection onto the first SVD plane. Blue line: 3.3s of data.
  \textbf{d, e, f)} Log predictive probability (blue) and entropy (green) over time for the respective datasets in (a,b,c). Black lines are exponential weighted moving averages of the data. Dashed green line indicates maximum entropy ($\log_2(N)$).} 
  \label{fig:exp}
\end{figure}

We next tested each algorithm on more complex data collected from neuroscience experiments. These data exhibited a variety of structure, from organized rotations (Figure \ref{fig:exp}a) to rapid transitions between noise clusters (Figure \ref{fig:exp}b) to slow dynamics (Figure \ref{fig:exp}c). In each case, Bubblewrap learned a tiling of the data that allowed it to equal or outperform state predictions from the comparison algorithms (Figure \ref{fig:exp}d--f, blue). In some cases, as with the mouse dataset, the algorithm of \cite{NIPS2016_b2531e7b} produced predictions for $x_t$ by degenerating to a random walk model (Table \ref{res-table} marked with *; Appendix \ref{app:additional-exp}). Regardless, Bubblewrap's tiling generated transition predictions with entropies far below those of a random walk (Figure \ref{fig:exp}d--f, green), indicating it successfully identified coarse structure, even in challenging datasets. Thus, even though these data are noise-dominated and lack much of the typical structure identified by neural population models, coarse-graining identifies some reliable patterns.

\begin{figure}
  \centering
  \includegraphics[width=\linewidth]{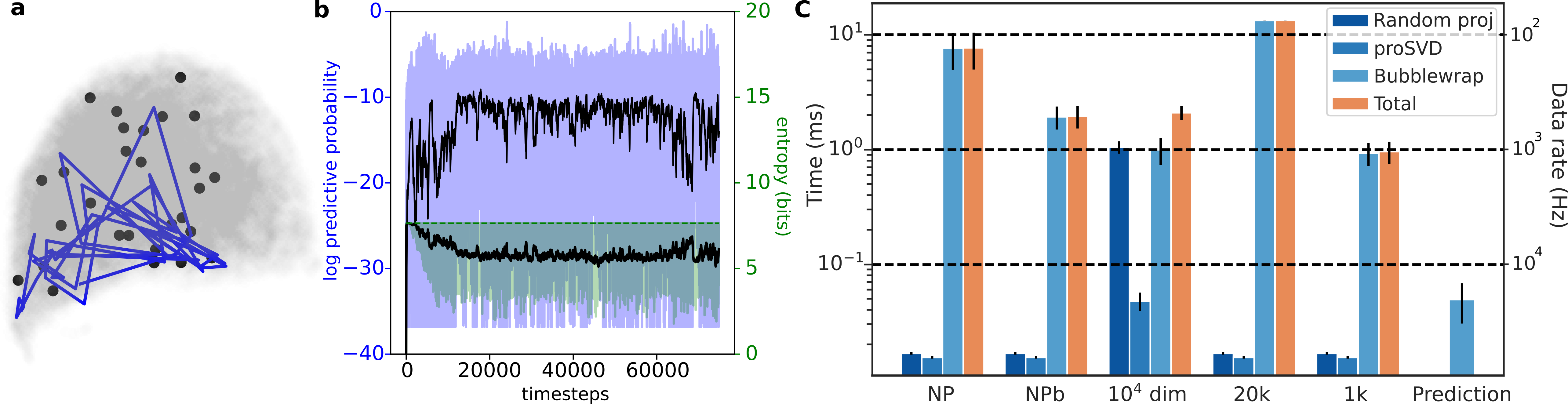}
  \caption{\textbf{High-throughput data \& benchmarking.} \textbf{a)} Bubblewrap results for example trajectories (blue) in the Neuropixels dataset \cite{steinmetz_pachitariu_stringer_carandini_harris_2019, stringer2019spontaneous} (data in gray) visualized with UMAP. \textbf{b)} Log predictive probability (blue) and entropy (green) over time. Black lines are exponential weighted moving averages of the data. Dashed green line indicates maximum entropy. \textbf{c)} Average cycle time (log scale) during learning or prediction (last bar) for each timepoint. Neuropixels (NP) is run as in (a,b) with no optimization and all heuristics, and Bubblewrap is easily able to learn at rates much faster than acquisition (30 ms). By turning off the global mean and covariance and priors updates and only taking a gradient step for $\mathcal{L}$ every 30 timepoints, we are able to run at close to 1 kHz (NPb). All other bars show example timings from Van der Pol synthetic datasets optimized for speed: $10^4$ dim, where we randomly project down to 200 dimensions and used proSVD to project to 10 dimensions for subsequent Bubblewrap modeling learning; $N$ = 20k, 10k, and 1k nodes, showing how our algorithm scales with the number of tiles; and Prediction, showing the time cost to predict one step ahead for the $N$ = 1k case. } 
  \label{fig:timing}
\end{figure}

We additionally considered the capability of our algorithm to scale to high-dimensional or high-sampling rate data. As a case study, we considered real-time processing (including random projections, proSVD, and Bubblewrap learning) of Neuropixels data comprising 2688 units with 74,330 timepoints from 30 ms bins. As Figure \ref{fig:timing} shows, Bubblewrap once again learns a tiling of the data manifold (a), capturing structure in the probability flow within the space (b) with predictive performance comparable to finer-grained methods (Table \ref{res-table}). More importantly, all these steps can be performed well within the 30ms per sample time of the data (c). In fact, when testing on representative examples of $d = 10^4$  dimensions, $1$ kHz sampling rates, or $N=20,000$ tiles, our algorithm was able to maintain amortized per-sample processing times below those of data acquisition. In practice, we found that even in higher-dimensional datasets (as in the Neuropixels case), only 1-2 thousand tiles were used by the model, making it easy to run at kHz data rates. What's more, while learning involved round trip GPU latencies to perform gradient updates, online predictions using slightly stale estimates of Bubblewrap parameters could be performed far faster, in tens of microseconds.

\begin{figure}[h!]
  \centering
  \includegraphics[width=\linewidth]{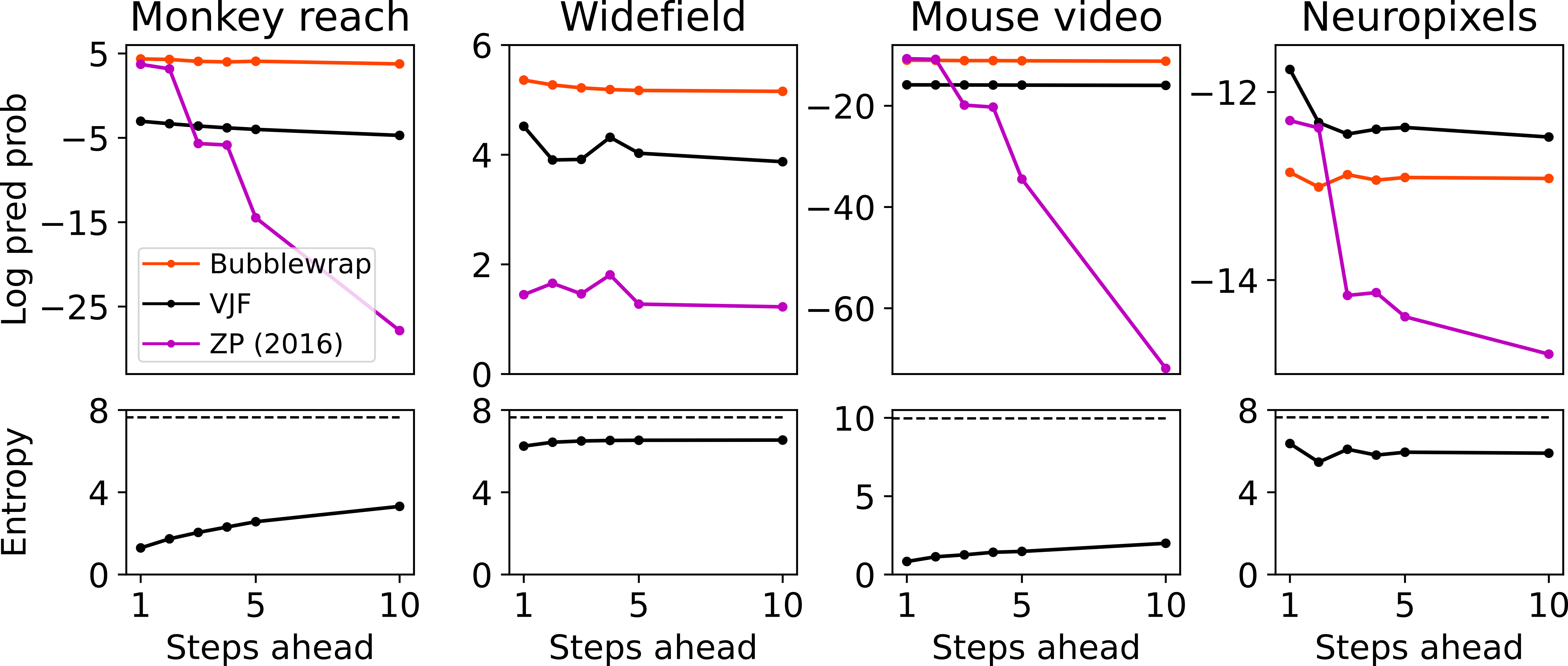}
  \caption{\textbf{Multi-step ahead predictive performance.} \textbf{(top)} Mean log predictive probability as a function of the number of steps ahead used for prediction for each of the four experimental datasets studied. Colors indicate model. \textbf{(bottom)} Bubblewrap entropy as a function of the number of steps ahead used for prediction. Higher entropy indicates more uncertainty about future states. Dashed lines denote maximum entropy for each dataset (log of the number of tiles). } 
  \label{fig:pred_ahead}
\end{figure}

Just as importantly, when used for closed loop experiments, algorithms must be able to produce predictions far enough into the future for interventions to be feasible. Thus we examined the performance of our algorithm and comparison models for predicting $T$ steps ahead into the future. Bubblewrap allows us to efficiently calculate predictions even many time steps into the future using (\ref{eqn:bubblewrap_pred}), whereas the comparison models require much costlier sampling approaches. Figure \ref{fig:pred_ahead} shows the mean log predictive probabilities for all models many steps into the future for each experimental dataset (top row), and the entropy of the predicted transitions using Bubblewrap (bottow row). 
Our algorithm consistently maintains performance even when predicting 10 steps ahead, providing crucial lead time to enable interventions at specific points along the learned trajectory. In comparison, predictive performance of \cite{NIPS2016_b2531e7b}, which initially matches or exceeds Bubblewrap for two datasets, rapidly declines, while Variational Joint Filtering \cite{zhao2020variational}, with lower log likelihood, also exhibits a slow decay in accuracy.

\section{Discussion}
While increasing attention has been paid in neuroscience to population hypotheses of neural function \cite{ebitz2021population}, and while many methods for modeling these data offline exist, surprisingly few methods function online, though presumably online methods will be needed to test some population dynamics hypotheses \cite{peixoto2021decoding}. While the neural engineering literature has long used online methods based on Kalman filtering, (e.g., \cite{bolus2020state}), and these methods are known to work well in many practical cases, they also imply strong assumptions about the evolution of activity within these systems. Thus, many studies that employ less constrained behavior or study neural activity with less robust dynamics may benefit from more flexible models that can be trained while the experiment is running. 

Here, to address this need, we have introduced both a new dimension reduction method that rapidly produces stable estimates of features and a method for rapidly mapping and charting transitions on neural manifolds. Rather than focus on moment-by-moment prediction, we focus on estimating a coarse tiling and probability flow among these tiles. Thus, Bubblewrap may be less accurate than methods based on dynamical systems when state trajectories are accurately described by smooth vector fields with Gaussian noise. Conversely, when noise dominates, is multimodal, or only large-scale probability flow is discernible over longer timescales, Bubblewrap is better poised to capture these features. We saw this in our experiments, where the model of \cite{NIPS2016_b2531e7b} exhibited better overall performance in the mouse video dataset (Figure \ref{fig:exp}e) when it did not learn to predict and degenerated to a random walk. Indeed, the most relevant comparison to the two approaches is the duality between stochastic differential equations and Fokker-Planck equations, where ours is a (softly) discretized analog of the latter. Nonetheless, in many of the cases we consider, Bubblewrap produces superior results even for state prediction. Nonetheless, like many similar models, ours includes multiple hyperparameters that require setting. While we did not experience catastrophic failure or sensitive dependence on parameters in our testing, and while our methods adapt to the scale and drift of the data, some tuning was required in practice.

As detailed above, while many methods target population dynamics, and a few target closed-loop settings \cite{yang2018control, bolus2020state,sani2020modeling}, very few models are capable of being trained online. Thus, the most closely related approaches are those in \cite{NIPS2016_b2531e7b,zhao2020variational}, to which we provide extensive comparisons. However, these comparisons are somewhat strained by the fact that we provided all models with the same proSVD-reduced low-dimensional data, while \cite{zhao2020variational} is capable of modeling high-dimensional data in its own right and \cite{NIPS2016_b2531e7b} was targeted at inferring neural computations from dynamical systems. We thus view this work as complementary to the dynamical systems approach, one that may be preferred when small distinctions among population dynamics are less important than characterizing highly noisy, non-repeating neural behavior.

Finally, we showed that online training of Bubblewrap can be performed fast enough for even kiloHertz data acquisition rates if small latencies are tolerable and gradient steps can be performed for small numbers of samples at a time. Yet, for real-time applications, it is not training time but the time required to make predictions that is relevant, and we demonstrate prediction times of tens of microseconds. 
Moreover, Bubblewrap is capable of producing effective predictions multiple time steps into the future, providing ample lead time for closed-loop interventions.
Thus, coarse-graining methods like ours open the door to online manipulation and steering of neural systems.

\begin{ack}
Research reported in this publication was supported by a NIH BRAIN Initiative Planning Grant (R34NS116738; JP), and a Swartz Foundation Postdoctoral Fellowship for Theory in Neuroscience (AD). AD also holds a Career Award at the Scientific Interface from the Burroughs Wellcome Fund.
\end{ack}

\bibliographystyle{unsrt}
\medskip
\bibliography{bubble}

\begin{thebibliography}{10}

\bibitem{ahrens2013whole}
Misha~B Ahrens, Michael~B Orger, Drew~N Robson, Jennifer~M Li, and Philipp~J
  Keller.
\newblock Whole-brain functional imaging at cellular resolution using
  light-sheet microscopy.
\newblock {\em Nature methods}, 10(5):413--420, 2013.

\bibitem{emiliani2015all}
Valentina Emiliani, Adam~E Cohen, Karl Deisseroth, and Michael H{\"a}usser.
\newblock All-optical interrogation of neural circuits.
\newblock {\em Journal of Neuroscience}, 35(41):13917--13926, 2015.

\bibitem{stevenson2011advances}
Ian~H Stevenson and Konrad~P Kording.
\newblock How advances in neural recording affect data analysis.
\newblock {\em Nature neuroscience}, 14(2):139--142, 2011.

\bibitem{steinmetz2018challenges}
Nicholas~A Steinmetz, Christof Koch, Kenneth~D Harris, and Matteo Carandini.
\newblock Challenges and opportunities for large-scale electrophysiology with
  neuropixels probes.
\newblock {\em Current opinion in neurobiology}, 50:92--100, 2018.

\bibitem{steinmetz2020neuropixels}
Nicholas~A Steinmetz, Cagatay Aydin, Anna Lebedeva, Michael Okun, Marius
  Pachitariu, Marius Bauza, Maxime Beau, Jai Bhagat, Claudia B{\"o}hm, Martijn
  Broux, et~al.
\newblock Neuropixels 2.0: A miniaturized high-density probe for stable,
  long-term brain recordings.
\newblock {\em bioRxiv}, 2020.

\bibitem{mante2013context}
Valerio Mante, David Sussillo, Krishna~V Shenoy, and William~T Newsome.
\newblock Context-dependent computation by recurrent dynamics in prefrontal
  cortex.
\newblock {\em nature}, 503(7474):78--84, 2013.

\bibitem{rajan2016recurrent}
Kanaka Rajan, Christopher~D Harvey, and David~W Tank.
\newblock Recurrent network models of sequence generation and memory.
\newblock {\em Neuron}, 90(1):128--142, 2016.

\bibitem{song2016training}
H~Francis Song, Guangyu~R Yang, and Xiao-Jing Wang.
\newblock Training excitatory-inhibitory recurrent neural networks for
  cognitive tasks: a simple and flexible framework.
\newblock {\em PLoS computational biology}, 12(2):e1004792, 2016.

\bibitem{cunningham2014dimensionality}
John~P Cunningham and M~Yu Byron.
\newblock Dimensionality reduction for large-scale neural recordings.
\newblock {\em Nature neuroscience}, 17(11):1500--1509, 2014.

\bibitem{ebitz2021population}
R~Becket Ebitz and Benjamin~Y Hayden.
\newblock The population doctrine revolution in cognitive neurophysiology.
\newblock {\em arXiv preprint arXiv:2104.00145}, 2021.

\bibitem{carlson2011sparse}
Eric~T Carlson, Russell~J Rasquinha, Kechen Zhang, and Charles~E Connor.
\newblock A sparse object coding scheme in area v4.
\newblock {\em Current Biology}, 21(4):288--293, 2011.

\bibitem{dimattina2013adaptive}
Christopher DiMattina and Kechen Zhang.
\newblock Adaptive stimulus optimization for sensory systems neuroscience.
\newblock {\em Frontiers in neural circuits}, 7:101, 2013.

\bibitem{NIPS2017_892c91e0}
Benjamin Cowley, Ryan Williamson, Katerina Clemens, Matthew Smith, and Byron~M
  Yu.
\newblock Adaptive stimulus selection for optimizing neural population
  responses.
\newblock In {\em Advances in Neural Information Processing Systems},
  volume~30, 2017.

\bibitem{abbasi2018deeptune}
Reza Abbasi-Asl, Yuansi Chen, Adam Bloniarz, Michael Oliver, Ben~DB Willmore,
  Jack~L Gallant, and Bin Yu.
\newblock The deeptune framework for modeling and characterizing neurons in
  visual cortex area v4.
\newblock {\em bioRxiv}, page 465534, 2018.

\bibitem{zhang2018closed}
Zihui Zhang, Lloyd~E Russell, Adam~M Packer, Oliver~M Gauld, and Michael
  H{\"a}usser.
\newblock Closed-loop all-optical interrogation of neural circuits in vivo.
\newblock {\em Nature methods}, 15(12):1037--1040, 2018.

\bibitem{bolus2020state}
Michael~F Bolus, Adam~A Willats, Christopher~J Rozell, and Garrett~B Stanley.
\newblock State-space optimal feedback control of optogenetically driven neural
  activity.
\newblock {\em Journal of Neural Engineering}, 2020.

\bibitem{peixoto2021decoding}
Diogo Peixoto, Jessica~R Verhein, Roozbeh Kiani, Jonathan~C Kao, Paul
  Nuyujukian, Chandramouli Chandrasekaran, Julian Brown, Sania Fong, Stephen~I
  Ryu, Krishna~V Shenoy, et~al.
\newblock Decoding and perturbing decision states in real time.
\newblock {\em Nature}, pages 1--6, 2021.

\bibitem{draelos2020online}
Anne Draelos and John Pearson.
\newblock Online neural connectivity estimation with noisy group testing.
\newblock {\em Advances in Neural Information Processing Systems}, 33, 2020.

\bibitem{brand2002incremental}
Matthew Brand.
\newblock Incremental singular value decomposition of uncertain data with
  missing values.
\newblock {\em Computer Vision?ECCV 2002}, pages 707--720, 2002.

\bibitem{baker2004block}
Christopher~G Baker.
\newblock A block incremental algorithm for computing dominant singular
  subspaces.
\newblock Master's thesis, Florida State University, 2004.

\bibitem{brand2006fast}
Matthew Brand.
\newblock Fast low-rank modifications of the thin singular value decomposition.
\newblock {\em Linear algebra and its applications}, 415(1):20--30, 2006.

\bibitem{baker2012low}
Christopher~G Baker, Kyle~A Gallivan, and Paul Van~Dooren.
\newblock Low-rank incremental methods for computing dominant singular
  subspaces.
\newblock {\em Linear Algebra and its Applications}, 436(8):2866--2888, 2012.

\bibitem{mairal2010online}
Julien Mairal, Francis Bach, Jean Ponce, and Guillermo Sapiro.
\newblock Online learning for matrix factorization and sparse coding.
\newblock {\em Journal of Machine Learning Research}, 11(1), 2010.

\bibitem{archer2015black}
Evan Archer, Il~Memming Park, Lars Buesing, John Cunningham, and Liam Paninski.
\newblock Black box variational inference for state space models.
\newblock {\em arXiv preprint arXiv:1511.07367}, 2015.

\bibitem{gao2016linear}
Yuanjun Gao, Evan Archer, Liam Paninski, and John~P Cunningham.
\newblock Linear dynamical neural population models through nonlinear
  embeddings.
\newblock {\em arXiv preprint arXiv:1605.08454}, 2016.

\bibitem{pandarinath2018inferring}
Chethan Pandarinath, Daniel~J O’Shea, Jasmine Collins, Rafal Jozefowicz,
  Sergey~D Stavisky, Jonathan~C Kao, Eric~M Trautmann, Matthew~T Kaufman,
  Stephen~I Ryu, Leigh~R Hochberg, et~al.
\newblock Inferring single-trial neural population dynamics using sequential
  auto-encoders.
\newblock {\em Nature methods}, 15(10):805--815, 2018.

\bibitem{linderman2017bayesian}
Scott Linderman, Matthew Johnson, Andrew Miller, Ryan Adams, David Blei, and
  Liam Paninski.
\newblock Bayesian learning and inference in recurrent switching linear
  dynamical systems.
\newblock In {\em Artificial Intelligence and Statistics}, pages 914--922,
  2017.

\bibitem{linderman2017recurrent}
Scott~W. Linderman*, Matthew~J. Johnson*, Andrew~C. Miller, Ryan~P. Adams,
  David~M. Blei, and Liam Paninski.
\newblock Bayesian learning and inference in recurrent switching linear
  dynamical systems.
\newblock In {\em Proceedings of the 20th International Conference on
  Artificial Intelligence and Statistics (AISTATS)}, 2017.

\bibitem{nassar2018tree}
Josue Nassar, Scott~W Linderman, Monica Bugallo, and Il~Memming Park.
\newblock Tree-structured recurrent switching linear dynamical systems for
  multi-scale modeling.
\newblock {\em arXiv preprint arXiv:1811.12386}, 2018.

\bibitem{NIPS2016_b2531e7b}
Yuan Zhao and Il~Memming Park.
\newblock Interpretable nonlinear dynamic modeling of neural trajectories.
\newblock In D.~Lee, M.~Sugiyama, U.~Luxburg, I.~Guyon, and R.~Garnett,
  editors, {\em Advances in Neural Information Processing Systems}, volume~29.
  Curran Associates, Inc., 2016.

\bibitem{yang2018control}
Yuxiao Yang, Allison~T Connolly, and Maryam~M Shanechi.
\newblock A control-theoretic system identification framework and a real-time
  closed-loop clinical simulation testbed for electrical brain stimulation.
\newblock {\em Journal of neural engineering}, 15(6):066007, 2018.

\bibitem{zhao2020variational}
Yuan Zhao and Il~Memming Park.
\newblock Variational online learning of neural dynamics.
\newblock {\em Frontiers in computational neuroscience}, 14, 2020.

\bibitem{gao2017theory}
Peiran Gao, Eric Trautmann, Byron Yu, Gopal Santhanam, Stephen Ryu, Krishna
  Shenoy, and Surya Ganguli.
\newblock A theory of multineuronal dimensionality, dynamics and measurement.
\newblock {\em BioRxiv}, page 214262, 2017.

\bibitem{trautmann2019accurate}
Eric~M Trautmann, Sergey~D Stavisky, Subhaneil Lahiri, Katherine~C Ames,
  Matthew~T Kaufman, Daniel~J O’Shea, Saurabh Vyas, Xulu Sun, Stephen~I Ryu,
  Surya Ganguli, et~al.
\newblock Accurate estimation of neural population dynamics without spike
  sorting.
\newblock {\em Neuron}, 103(2):292--308, 2019.

\bibitem{achlioptas2003database}
Dimitris Achlioptas.
\newblock Database-friendly random projections: Johnson-lindenstrauss with
  binary coins.
\newblock {\em Journal of computer and System Sciences}, 66(4):671--687, 2003.

\bibitem{li2006very}
Ping Li, Trevor~J Hastie, and Kenneth~W Church.
\newblock Very sparse random projections.
\newblock In {\em Proceedings of the 12th ACM SIGKDD international conference
  on Knowledge discovery and data mining}, pages 287--296, 2006.

\bibitem{johnson1984extensions}
William~B Johnson and Joram Lindenstrauss.
\newblock Extensions of lipschitz mappings into a hilbert space.
\newblock {\em Contemporary mathematics}, 26(189-206):1, 1984.

\bibitem{schonemann1966generalized}
Peter~H Sch{\"o}nemann.
\newblock A generalized solution of the orthogonal procrustes problem.
\newblock {\em Psychometrika}, 31(1):1--10, 1966.

\bibitem{degenhart2020stabilization}
Alan~D Degenhart, William~E Bishop, Emily~R Oby, Elizabeth~C Tyler-Kabara,
  Steven~M Chase, Aaron~P Batista, and M~Yu Byron.
\newblock Stabilization of a brain--computer interface via the alignment of
  low-dimensional spaces of neural activity.
\newblock {\em Nature biomedical engineering}, 4(7), 2020.

\bibitem{ross2008incremental}
David~A Ross, Jongwoo Lim, Ruei-Sung Lin, and Ming-Hsuan Yang.
\newblock Incremental learning for robust visual tracking.
\newblock {\em International journal of computer vision}, 77(1):125--141, 2008.

\bibitem{berman2014mapping}
Gordon~J Berman, Daniel~M Choi, William Bialek, and Joshua~W Shaevitz.
\newblock Mapping the stereotyped behaviour of freely moving fruit flies.
\newblock {\em Journal of The Royal Society Interface}, 11(99):20140672, 2014.

\bibitem{berman2016predictability}
Gordon~J Berman, William Bialek, and Joshua~W Shaevitz.
\newblock Predictability and hierarchy in drosophila behavior.
\newblock {\em Proceedings of the National Academy of Sciences},
  113(42):11943--11948, 2016.

\bibitem{pereira2020quantifying}
Talmo~D Pereira, Joshua~W Shaevitz, and Mala Murthy.
\newblock Quantifying behavior to understand the brain.
\newblock {\em Nature Neuroscience}, pages 1--13, 2020.

\bibitem{mongillo2008online}
Gianluigi Mongillo and Sophie Deneve.
\newblock Online learning with hidden markov models.
\newblock {\em Neural computation}, 20(7):1706--1716, 2008.

\bibitem{cappe2009line}
Olivier Capp{\'e} and Eric Moulines.
\newblock On-line expectation--maximization algorithm for latent data models.
\newblock {\em Journal of the Royal Statistical Society: Series B (Statistical
  Methodology)}, 71(3):593--613, 2009.

\bibitem{le2013online}
Sylvain Le~Corff, Gersende Fort, et~al.
\newblock Online expectation maximization based algorithms for inference in
  hidden markov models.
\newblock {\em Electronic Journal of Statistics}, 7:763--792, 2013.

\bibitem{kingma2014adam}
Diederik~P Kingma and Jimmy Ba.
\newblock Adam: A method for stochastic optimization.
\newblock {\em arXiv preprint arXiv:1412.6980}, 2014.

\bibitem{churchlandwebsite}
Mark Churchland.
\newblock Churchland lab code.
\newblock
  \url{"https://churchland.zuckermaninstitute.columbia.edu/content/code"}.

\bibitem{churchland2012neural}
Mark~M Churchland, John~P Cunningham, Matthew~T Kaufman, Justin~D Foster, Paul
  Nuyujukian, Stephen~I Ryu, and Krishna~V Shenoy.
\newblock Neural population dynamics during reaching.
\newblock {\em Nature}, 487(7405):51--56, 2012.

\bibitem{musall2019single}
Simon Musall, Matthew~T Kaufman, Ashley~L Juavinett, Steven Gluf, and Anne~K
  Churchland.
\newblock Single-trial neural dynamics are dominated by richly varied
  movements.
\newblock {\em Nature neuroscience}, 22(10):1677--1686, 2019.

\bibitem{musalldata}
Simon Musall, Matthew~T. Kaufman, Ashley~L. Juavinett, Steven Gluf, and Anne~K.
  Churchland.
\newblock Single-trial neural dynamics are dominated by richly varied
  movements: dataset.
\newblock Technical report, 2019.

\bibitem{steinmetz_pachitariu_stringer_carandini_harris_2019}
Nick Steinmetz, Marius Pachitariu, Carsen Stringer, Matteo Carandini, and
  Kenneth Harris.
\newblock Eight-probe neuropixels recordings during spontaneous behaviors, Mar
  2019.

\bibitem{stringer2019spontaneous}
Carsen Stringer, Marius Pachitariu, Nicholas Steinmetz, Charu~Bai Reddy, Matteo
  Carandini, and Kenneth~D Harris.
\newblock Spontaneous behaviors drive multidimensional, brainwide activity.
\newblock {\em Science}, 364(6437), 2019.

\bibitem{jax2018github}
James Bradbury, Roy Frostig, Peter Hawkins, Matthew~James Johnson, Chris Leary,
  Dougal Maclaurin, George Necula, Adam Paszke, Jake Vander{P}las, Skye
  Wanderman-{M}ilne, and Qiao Zhang.
\newblock {JAX}: composable transformations of {P}ython+{N}um{P}y programs.
\newblock 2018.

\bibitem{sani2020modeling}
Omid~G Sani, Hamidreza Abbaspourazad, Yan~T Wong, Bijan Pesaran, and Maryam~M
  Shanechi.
\newblock Modeling behaviorally relevant neural dynamics enabled by
  preferential subspace identification.
\newblock Technical report, Nature Publishing Group, 2020.

\end{thebibliography}

\newpage
\setcounter{section}{0}
\renewcommand{\thesection}{\Alph{section}}
\setcounter{figure}{0}
\renewcommand{\thefigure}{S\arabic{figure}}
\pagenumbering{arabic}

\appendix

\section{Details of proSVD algorithm}
\label{app:prosvd-details}
We follow the notation of \cite{baker2004block,baker2012low} with the exception that we use $X$ for the data matrix rather than $A$, with new data $X_+$ of dimension $n \times b$ rather than $m \times l$. Here, we focus on the updates of the left singular subspace spanned. Details for the right singular subspace are similar and covered in the original works. Matrix sizes are listed for convenience in Table \ref{tab:prosvd-dims}.

\begin{table}[h]
  \centering
  \caption{Matrix dimensions for incremental SVD.}
  \begin{tabular}{ccc}
      matrix & rows & columns \\
      \toprule
      $X$ & $n$ & $T$ \\
      $X_0$ & $n$ & $k$ \\
      $X_+$ & $n$ & $b$ \\
      $C$ & $k$ & $b$ \\
      $Q_t$ & $n$ & $k$ \\
      $R_t$ & $k$ & $k$ \\
      $\hat{Q}$ & $n$ & $k + b$\\
      $\hat{R}$ & $k + b$ & $k + b$ \\
      $U$ & $k + b$ & $k + b$ \\
      $U_1$ & $k + b$ & $k$ \\
      $U_2$ & $k + b$ & $b$ \\
      $T_u$ & $k$ & $k$ \\
      $S_u$ & $b$ & $b$ \\
      $G_u$ & $k + b$ & $k + b$ \\
      $G_{u_1}$ & $k + b$ & $k$ \\
  \end{tabular}
  \label{tab:prosvd-dims}
\end{table}
The goal of the algorithm is to maintain an approximation of the data up to the present moment as
\begin{equation}
  \label{qrw_fact}
  X \approx Q R W^\top
\end{equation}
with $Q$ and $W$ orthogonal but $R$ not necessarily diagonal. The requirement that (\ref{qrw_fact}) be equivalent to the top-$k$ SVD requires that the top-$k$ SVD of $X$ can easily be computed via the SVD of the small matrix $R = U \Sigma V^\top$.

The algorithm begins with an initial data matrix $X_0$, which is factorized via the QR decomposition
\begin{equation}
  X_0 = \begin{bmatrix}
    Q_0 & Q_\perp
  \end{bmatrix} 
  \begin{bmatrix}
    R_0 \\ 0
  \end{bmatrix}
  \mathbbm{1} ,
\end{equation}
which has the form (\ref{qrw_fact}) if we identify $W_0 = \mathbbm{1}$. Thus $Q_0$ forms the initial candidate for a basis for the top-$k$ subspace. On subsequent iterations, the procedure is as follows:
\begin{enumerate}
  \item Observe a new $n\times b$ data matrix $X_+$.
  \item Perform a Gram-Schmidt Orthogonalization of this new data, obtaining $C$, a projection into the previous basis $Q_{t-1}$ and a remainder, $X_\perp$.
  \item Perform a QR decomposition $X_\perp = Q_\perp R_\perp$. This gives rise to a new factorization
    \begin{equation}
      X_t = 
      \begin{bmatrix}
        X_{t-1} & X_+
      \end{bmatrix}
      = \hat{Q} \hat{R} \hat{W}^T 
    \end{equation}
    with 
    \begin{align}
      \hat{Q} &\equiv \begin{bmatrix} Q_{t-1} & Q_\perp \end{bmatrix} \\
      \hat{R} &\equiv \begin{bmatrix} R_{t-1} & C \\ 0 & R_\perp \end{bmatrix} .
    \end{align}
  \item From this new factorization, the goal is to block diagonalize $\hat{R}$ and truncate to the upper left block, which is the new top-$k$ singular subspace. This is done by first performing an SVD, $\hat{R} = U \Sigma V^\top$.
  \item To allow for old and stale data to decay in influence over time and to prevent unbounded accumulation of variance in $\Sigma$, we follow \cite{ross2008incremental} in multiplying $\Sigma$ by a discount parameter $\alpha$ at each step.
  \item As in \cite{baker2004block,baker2012low}, the goal is to find a matrix $G_u$ (and a counterpart on the right, $G_v$) satisfying 
    \begin{equation}
      G_u^\top U = \begin{bmatrix}
        T_u & 0 \\
        0 & S_u
      \end{bmatrix}
    \end{equation}
    or the equivalent condition 
    \begin{equation}
      \label{G_condition}
      G_u^\top U_1 = \begin{bmatrix}
        T_u \\ 
        0
      \end{bmatrix} 
    \end{equation}
    with $U_1$ the first $k$ columns of $U$, 
    which would yield 
    \begin{equation}
      G_u^\top \hat{R} G_v = \begin{bmatrix}
        T_u \Sigma_1 T_v^\top & 0 \\
        0 & S_u \Sigma_2 S_v^\top
      \end{bmatrix} .
    \end{equation}
\item This done, the matrices can once again be truncated back to their top-$k$ versions:
  \begin{align}
    R_t &= T_u \Sigma_1 T_v^\top \\     
    Q_t &= \hat{Q}G_{u_1} = \hat{Q}U_1 T_u^\top .\label{Q_update}
  \end{align}
\end{enumerate}

What is most important to note in this is that the solution to (\ref{G_condition}) is not unique. Many choices of $G_u$ (equivalently $T_u$) are possible. In \cite{baker2004block,baker2012low}, a particular solution to this equation is chosen for computational efficiency. By contrast, proSVD seeks to solve
\begin{equation}
  \min_{T_u} \lVert Q_t - Q_{t-1}\rVert_F = \min_{T_u} \lVert \hat{Q}U_1 T_u^\top - Q_{t-1}\rVert_F,
\end{equation}
with $\lVert \cdot \rVert_F$ denoting the Frobenius norm. As previously stated, this is an Orthogonal Procrustes problem with solution $T = \tilde{U}\tilde{V}^\top$, where $\tilde{U}$ and $\tilde{V}$ are defined by the SVD \cite{schonemann1966generalized}:
\begin{equation}
  \label{op_step}
  \tilde{U} \tilde{\Sigma} \tilde{V}^\top = Q_{t-1}^\top \hat{Q} U_1 = \begin{bmatrix}
    \mathbbm{1}_{k \times k} & \mathbf{0}_{k \times b}
  \end{bmatrix} U_1 .
\end{equation}
That is, $T$ can be found by performing the SVD of the upper left $k \times k$ block of $U$, and the only additional cost of our formulation relative to \cite{baker2004block,baker2012low} is the $\mathcal{O}(k^3)$ cost of this SVD, which is negligible in practice for $k$ small (Figure \ref{fig:prosvd}b).


\section{Details of Bubblewrap initialization and heuristics}
\label{app:bubblewrap_details}
Our implementation of Bubblewrap neither normalizes nor assumes a scale for incoming data. Thus, in order to choose priors that adjust to the scale of the data, we adopt an empirical Bayes approach. That is, we adjust some parameters of the Normal-Inverse-Wishart priors for each bubble, $\mu_{0j}$ and $\Psi_j$, based on data. More specifically, we first calculate empirical estimates $\bar{\mu}$ and $\overline{\Sigma}$ of the data mean and covariance online, as in line 11 of Algorithm \ref{alg:bubblewrap}. We then update the priors as follows:

\textbf{Random walk dynamics on $\mu_{0j}$:} If $\mu_{0j}$ is fixed and identical across all bubbles, then points passing through this concentration of Gaussians will be assigned equally to all, which creates problems for tiling. Here, the idea is to allow the $\mu_{0j}$ to randomly walk, thereby breaking degeneracy among nodes with no data points. However, in order to bound the random walk, we include a decay back toward the current center of mass. More specifically, each of the $\mu_{0j}$ is governed by a biased random walk:
\begin{equation}
    \label{eqn:mu0_drift}
    \mu_{0j}(t) = (1 - \lambda)\mu_{0j}(t - 1) + \lambda \bar{\mu}(t - 1) +  \epsilon_{jt}
\end{equation}
with $\lambda \in [0, 1]$ and $\epsilon_{jt} \sim \mathcal{N}(0, \eta^2)$, $\mathbb{E}[\epsilon_{jt}\epsilon_{j't'}] = \eta^2\delta_{jj'}\delta_{tt'}$. Here, $\eta$ is a step size for the random walk that may differ by dimension. Note that, when $\bar{\mu}$ is fixed, the distribution of $\mu_{0j}$ converges to $\mathcal{N}(\bar{\mu}, \mathrm{diag}(\eta^2)/\lambda)$. 

In practice, we chose $\lambda = 0.02$ and $\eta = \sqrt{\lambda \; \text{diag}(\overline{\Sigma})}$, so that the diffusion range of bubbles with no data is set by the scale of the data distribution. 

\textbf{Bubble packing for $\Psi_j$:} Here, the basic idea is to choose $\Psi_j$, which controls the covariance around which the prior concentrates, so that the total volume of bubbles scales as the volume of the data. Specifically, let $\Psi_j = \sigma^2 \mathbbm{1}$, so that the covariance prior is spherical with radius $\sigma$. We also assume that, in $k$ dimensions, data take up $\mathcal{O}(L)$ space along each dimension, so that $\mathrm{vol}(data) \sim L^k$. Then the volume of each bubble is $\sim \sigma^k$ and the total volume of all bubbles is
\begin{equation}
    N \sigma^k \sim L^k \quad \Rightarrow \quad \sigma \sim \frac{L}{N^{\frac{1}{k}}}
\end{equation}
which yields
\begin{equation}
    \Psi_j = \frac{L^2}{N^{\frac{2}{k}}} \mathbbm{1}.
\end{equation}
More generally, we want $\Psi_j$ to reflect the estimated data covariance $\overline{\Sigma}$, so we set
\begin{equation}
    \Psi_j = \frac{1}{N^{\frac{2}{k}}}\overline{\Sigma} .
\end{equation}

\section{Predictive distribution calculations}
\label{app:predictive_calcs}
Both \cite{NIPS2016_b2531e7b} and \cite{zhao2020variational} use a nonlinear dynamical system model parameterized as
\begin{align}
  \mathbf{x}_{t+1} &= \mathbf{x}_t + f(\mathbf{x}_t) + \mathbf{B}_t(\mathbf{x}_t) \mathbf{u}_t + \epsilon_{t + 1} \label{eqn:zp_mean_pred}\\
  \mathbf{y}_t &\sim P(g(\mathbf{Cx}_t + \mathbf{b})) \label{eqn:zp_obs_model}\\
  f(\mathbf{x}) &= \mathbf{W}\boldsymbol{\phi}(\mathbf{x}) - e^{-\tau^2}\mathbf{x}_t\\
  \mathbf{x}_0, \epsilon_t &\sim \mathcal{N}(\mathbf{0}, \sigma^2 \mathbbm{1}) ,
\end{align}
with $\boldsymbol{\phi}$ a set of basis functions and $g$ a link function. In \cite{zhao2020variational}, $\mathbf{B}$ is assumed constant and $\tau \rightarrow \infty$. More importantly, the filtered probability $p(\mathbf{x}_t|\mathbf{y}_{\le t})$ is approximated by a posterior $q(\mathbf{x}_t) = \mathcal{N}(\boldsymbol{\mu}_t, \mathrm{diag}(\mathbf{s}_t))$ with $\boldsymbol{\mu}_t$ and $\mathbf{s}_t$ defined by the output of a neural network trained to optimize a variational lower bound on the log evidence.

In \cite{NIPS2016_b2531e7b}, the loss function is mean squared error in the prediction (\ref{eqn:zp_mean_pred}), which is equivalent to the likelihood model
\begin{equation}
  \label{eqn:zp_2016:pred_likelihood}
  p(\mathbf{x}_{t+1}|\mathbf{x}_t) = \mathcal{N}(\mathbf{x}_t + f(\mathbf{x}_t) + \mathbf{B}_t(\mathbf{x}_t) \mathbf{u}_t, \sigma^2 \mathbbm{1}), 
\end{equation}
with $\sigma^2 = \mathrm{var}[\epsilon_t]$ estimated from the data. In \cite{NIPS2016_b2531e7b}, there is no separate observation model (\ref{eqn:zp_obs_model}), so $\mathbf{y} = \mathbf{x}$, and we use (\ref{eqn:zp_2016:pred_likelihood}) (with $\sigma^2$ estimated from the residuals of (\ref{eqn:zp_mean_pred}) via an exponential smooth) to calculate predictive likelihood for our comparisons.

For \cite{zhao2020variational}, following the procedure outlined there, we can compute the log predictive probability via sampling
\begin{align}
  \mathbf{x}^s_t &\sim q(\mathbf{x}_t) \\
  \mathbf{x}^s_{t+1} &\sim \mathcal{N}(\mathbf{x}^s_t + f(\mathbf{x}^s_t) + \mathbf{B}_t \mathbf{u}_t, \sigma^2) \\
  \log p(\mathbf{y}_{t+1}|\mathbf{y}_{\le t}) &\approx \log \left[\frac{1}{S}\sum_{s=1}^S P(\mathbf{y}_{t+1} ;g(\mathbf{Cx}^s_{t+1} + \mathbf{b})) \right]
\end{align}
with $S = 100$ samples. However, for direct comparison with our method, which operates in a reduced-dimensional space, we compared the predictive probability by feeding all methods the same data as reduced by proSVD, so that $\mathrm{dim}(\mathbf{y}) = \mathrm{dim}(\mathbf{x}) = \mathrm{dim}(\mathrm{data})$.

When predicting more than one time step ahead, we use sequential sampling for both models. For the model of \cite{NIPS2016_b2531e7b}, we iterate (\ref{eqn:zp_mean_pred}) to produce $S$ trajectories $T-1$ time steps into the future (ending at $x^s_{t+T-1}$ and calculate
\begin{equation}
  p(\mathbf{x}_{t+T}|\mathbf{x}_{\le t}) = \frac{1}{S}\sum_{s=1}^S p(\mathbf{x}_{t+T}|\mathbf{x}^s_{t + T -1}) = \frac{1}{S}\sum_{s=1}^S\mathcal{N}(\mathbf{x}^s_{t+T-1} + f(\mathbf{x}^s_{t+T-1}) , \sigma^2 \mathbbm{1})
\end{equation}
using $S=100$ trajectories. For the model of \cite{zhao2020variational}, we have 
\begin{align}
  p(\mathbf{y}_{t+T}|\mathbf{y}_{\le t}) &= \int \! \prod_{t'=t+1}^T \!\! d\mathbf{x}_{t'} \; p(\mathbf{y}_{t+T}|\mathbf{x}_{t+1:T}) p(\mathbf{x}_{t+1:T}|\mathbf{x}_{t}) p(\mathbf{x}_{t}|\mathbf{y}_{\le t}) \nonumber \\
  &\approx \int \! \prod_{t'=t+1}^T \!\! d\mathbf{x}_{t'} \; p(\mathbf{y}_{t + T}|\mathbf{x}_{t+1:T}) p(\mathbf{x}_{t+1:T}|\mathbf{x}_{t}) q(\mathbf{x}_{t}) \nonumber \\
  &\approx \frac{1}{S}\sum_{s=1:S} p(\mathbf{y}_{t+T}|\mathbf{x}^{s}_{t+T})  \nonumber \\
  &= \frac{1}{S}\sum_{s=1}^S P(\mathbf{y}_{t+T} ;g(\mathbf{Cx}^s_{t+T} + \mathbf{b})),
\end{align}
where again we have used sampling to marginalize over the intervening latent states.



\section{Additional Experiments}
\label{app:additional-exp}

All experimental simulations were run on custom-built desktop machines running either Ubuntu 18.04.4 LT or Ubuntu 20.04.2. The computers ran with either 64GB or 128 GB of system memory, and various CPUs including a 4 or 8 core 4.0 GHz Intel i7-6700K processor and 14 core 3.1 GHz Intel i9-7940X processors. GPUs used included an NVIDIA GeForce GTX 1080 Ti (11 GB), 2x NVIDIA RTX 2080 Ti (11 GB), an NVIDIA Titan Xp (12 GB), and an NVIDIA RTX 3090 (24 GB).

\subsection{Datasets}
The monkey reach dataset can be found at \url{https://churchland.zuckermaninstitute.columbia.edu/content/code} \cite{churchlandwebsite}, and was originally presented in \cite{churchland2012neural}. The widefield calcium imaging and mouse video datasets can both be found at \url{https://dx.doi.org/10.14224/1.38599} \cite{musalldata}, and were originally presented in \cite{musall2019single}. The neuropixels dataset can be found at \url{https://doi.org/10.25378/janelia.7739750.v4} \cite{steinmetz_pachitariu_stringer_carandini_harris_2019}, and has a (CC BY-NC 4.0) license. It was originally presented in \cite{stringer2019spontaneous}. 

\begin{figure}[h!]
    \centering
    \includegraphics[width=\textwidth]{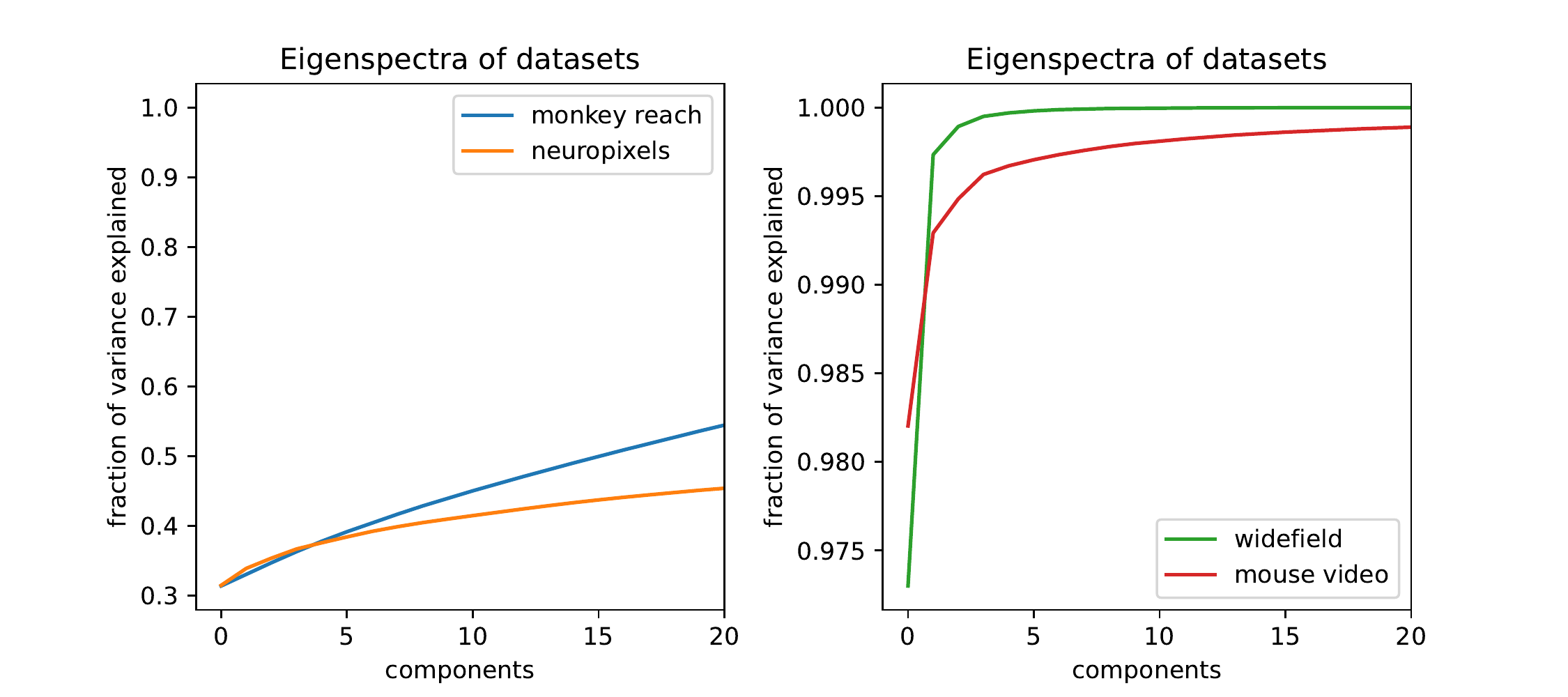}
    \caption{\textbf{Eigenspectra of datasets.} Cumulative variance explained by retaining differing numbers of components in linear dimensionality reduction. Labels reference datasets as in the main text.}
    \label{fig:spectrafig}
\end{figure}

To motivate the number of dimensions of the low-dimensional subspace onto which we project the data, we show the cumulative variance explained by each eigenvector of the real datasets in Figure \ref{fig:spectrafig}. These spectra show that keeping on the order of $\sim 10$ components retains $40\%$ of variance of the monkey reach and neuropixels datasets, and retains $>99\%$ variance of the widefield and mouse video datasets. 

\subsection{proSVD stability}
\begin{figure}[h!]
    \centering
    \includegraphics[width=0.75\textwidth]{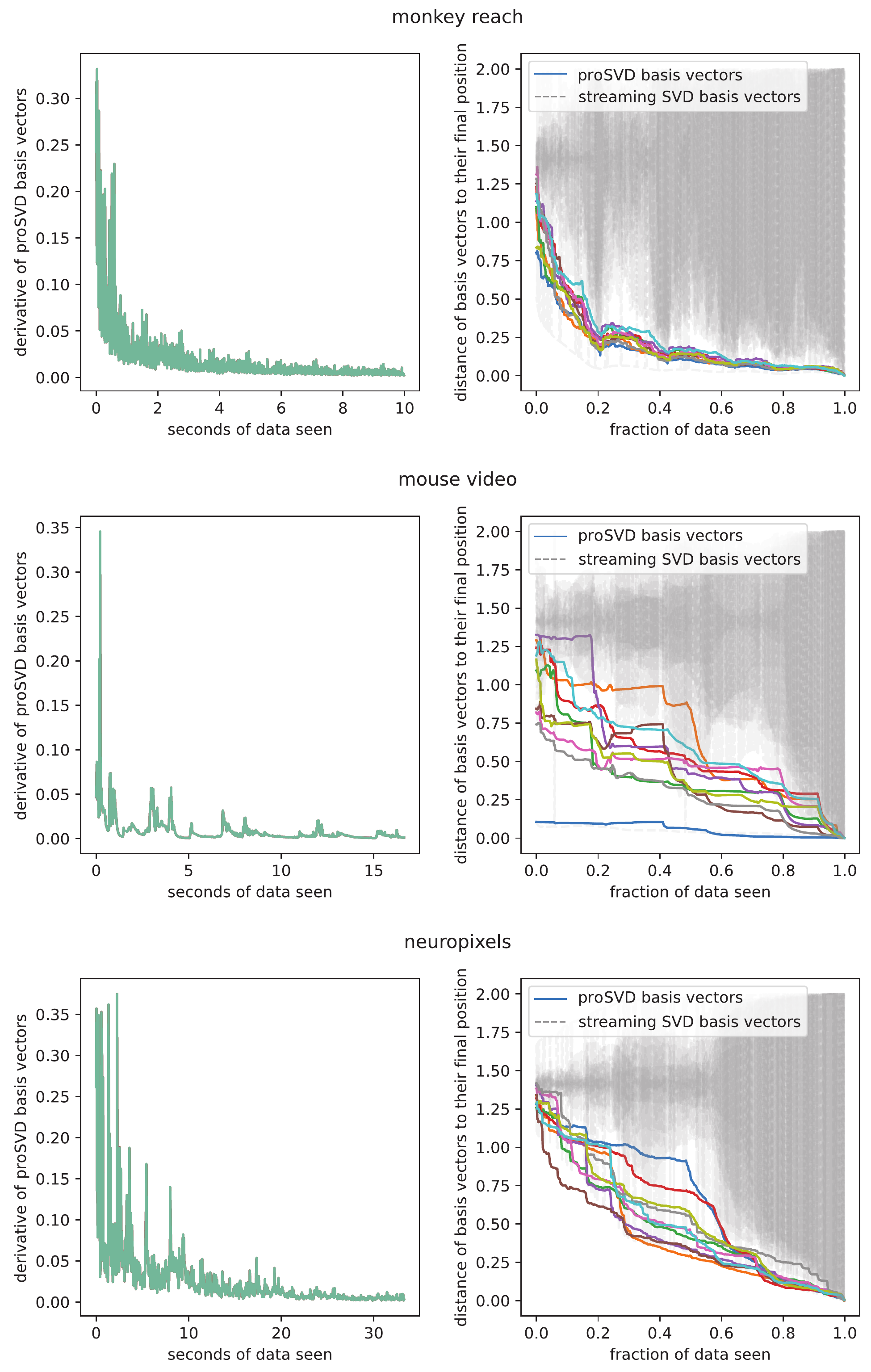}
    \caption{\textbf{Rapid stability of proSVD.} Titles reference data as in the main text. First column shows the derivatives of the proSVD basis vectors over time, i.e., the norms of the first difference of the basis vectors learned over time. Derivatives of streaming SVD vectors not shown for clarity. Second column shows how close the streaming vectors are to their final positions, measured by the Euclidean distance from the whole data basis vectors to the most recently learned basis vectors. $k=10$ vectors shown.}
    \label{fig:suppstability}
\end{figure}

Here we apply proSVD and the incremental block update method of \cite{baker2004block, baker2012low} (referred to as streaming SVD) to the experimental datasets used in the main text. The first column of Figure \ref{fig:suppstability} shows that the proSVD basis vectors stabilize after a few tens of seconds for these datasets, with only small gradients thereafter. The second column shows that proSVD basis vectors stabilize to their final (offline) positions faster than streaming SVD vectors. 

\subsection{Log predictive probability \& entropy}

\begin{figure}
  \centering
  \includegraphics[width=\linewidth]{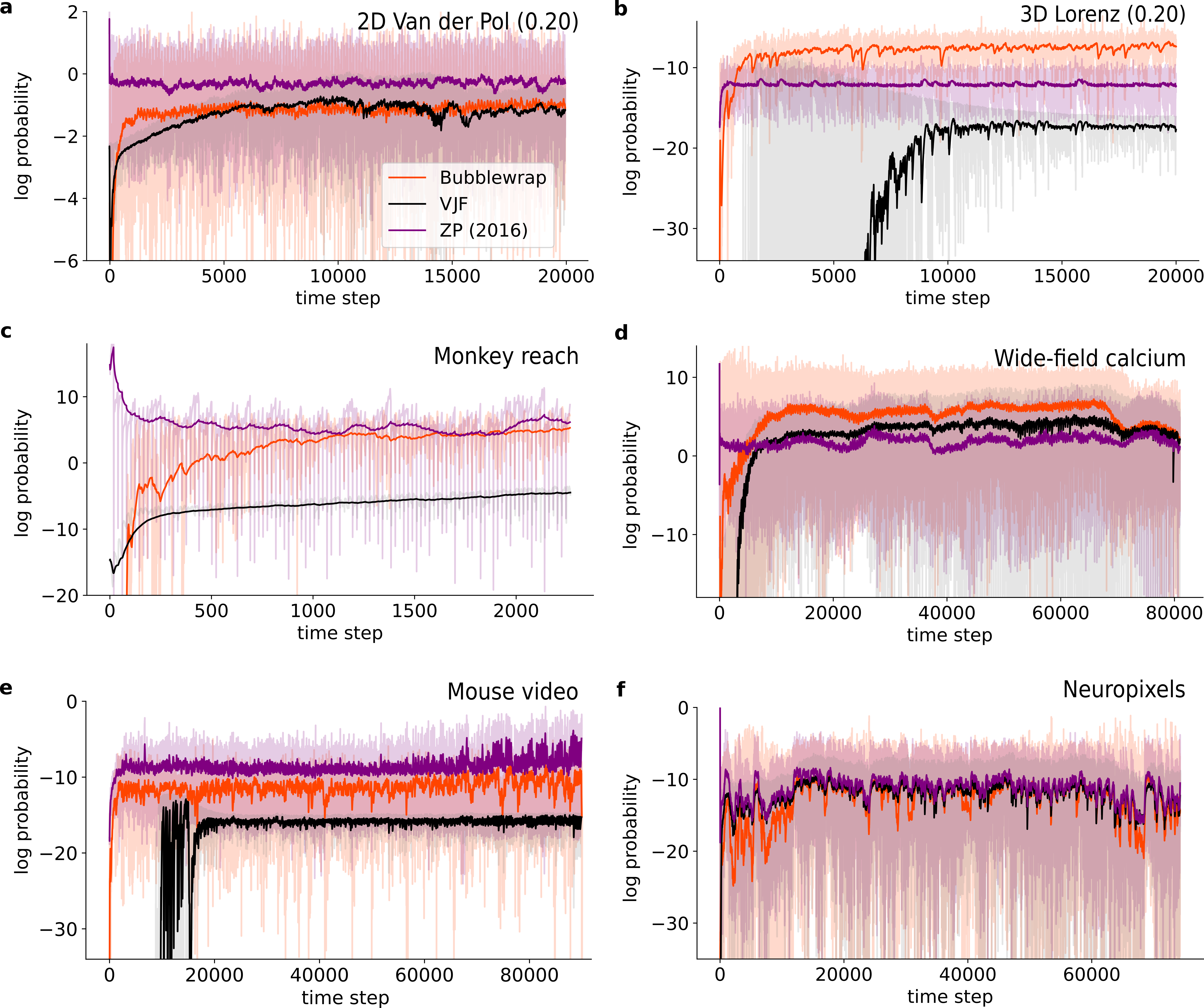}
  \caption{\textbf{Log predictive probability results} for all systems not shown in the main text. Log predictive probability across all timepoints for \textbf{a)} 2D Van der Pol (0.20 noise), \textbf{b)} 3D Lorenz (0.20 noise), \textbf{c)} Monkey reach, \textbf{d)} Wide-field calcium, \textbf{e)} Mouse video, and \textbf{f)} Neuropixels datasets. Conventions are as in Figure \ref{fig:2d3d}.}
  \label{fig:supplogfig}
\end{figure}

\begin{figure}
  \centering
  \includegraphics[width=0.75\linewidth]{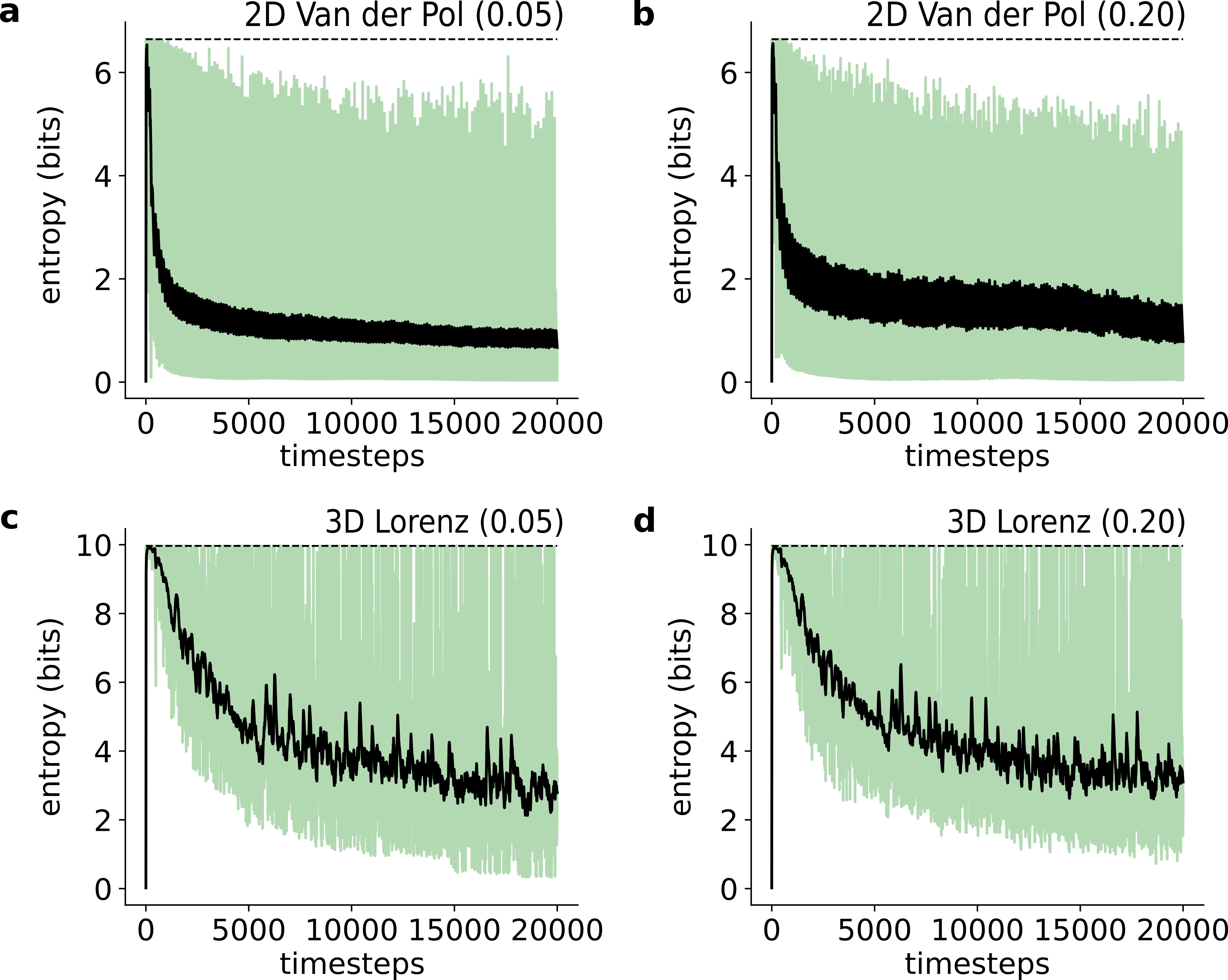}
  \caption{\textbf{Entropy} across all timepoints for each comparative model for all systems not shown in main text. Black lines are exponential weighted moving averages of the data. \textbf{a)} 2D Van der Pol (0.05), \textbf{b)} 2D Van der Pol (0.20), \textbf{c)} 3D Lorenz (0.05), \textbf{d)} and 3D Lorenz (0.20) datasets.}
  \label{fig:suppentfig}
\end{figure}

Here we include results for log predictive probability and entropy over time for the datasets not included in the main text. Figure \ref{fig:supplogfig} shows the log predictive probability across all time points for all datasets except the 2D Van der Pol and 3D Lorenz (5 \% noise) datasets (cp. Figure \ref{fig:2d3d}). Figure \ref{fig:suppentfig} plots the entropy over time for the 2D Van der Pol (0.05, 0.20) and the 3D Lorenz (0.05, 0.20) datasets ; all other entropy results were shown in the main text (Figure \ref{fig:exp}d-f, Figure \ref{fig:timing}b). 

\subsection{Final transition matrices}

\begin{figure}[h]
    \centering
    \includegraphics[width=0.55\textwidth]{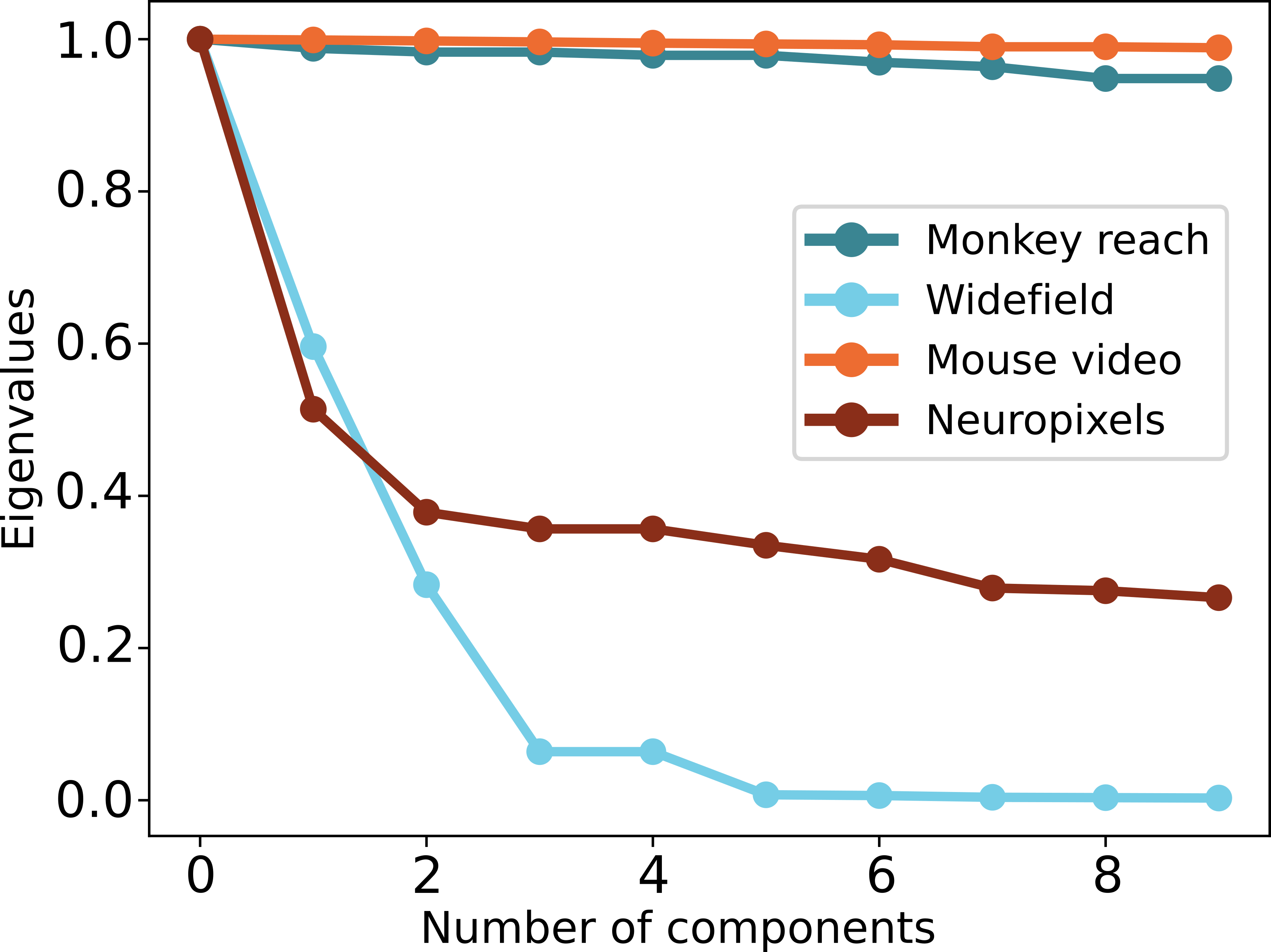}
    \caption{\textbf{Eigenspectra of final learned transition matrices.} Labels reference datasets as in the main text. Both the mouse video and monkey reach datasets have many eigenvalues close to 1, implying slow mixing of the corresponding Markov chain and long-range predictive power. By contrast, the widefield calcium and neuropixels datasets, with faster-decaying eigenspectra, quickly lose predictive information over time.}
    \label{fig:suppspectrafig}
\end{figure}

Figure \ref{fig:suppspectrafig} shows the eigenspectra of the final learned transition matrices $A$ for each of the four experimental datasets studied in the main text. The monkey reach and mouse video results yielded slowly-decaying eigenspectra, suggesting long-range predictive power, whereas the widefield and neuropixels eigenspectra decayed much faster (and thus yielded less long-range predictive power over time). Figures \ref{fig:exp} and \ref{fig:timing} in the main text also show this distinction: the monkey reach and mouse video entropy results reached much lower values than the other two datasets, corresponding to lower-uncertainty predictions. 

\begin{figure}
    \centering
    \includegraphics[width=0.75\textwidth]{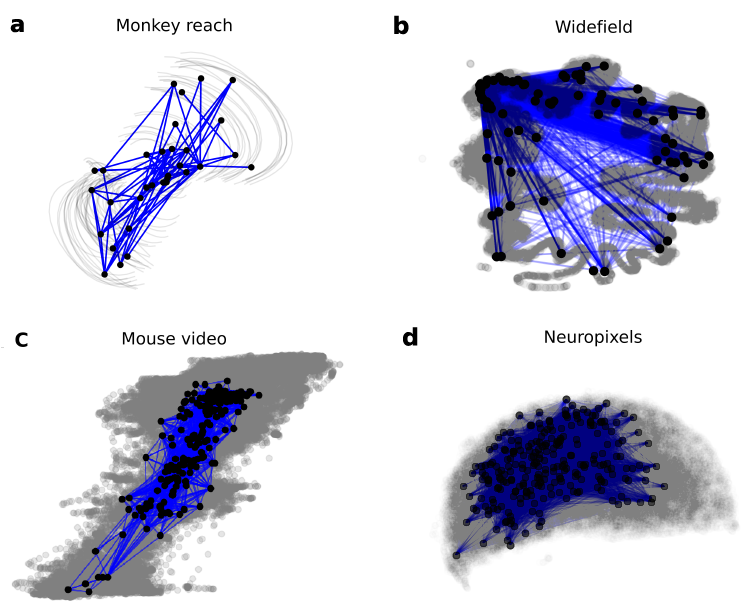}
    \caption{\textbf{Final transition matrices for each experimental dataset.} Labels reference datasets as in the main text. Data are plotted in grey as trajectories \textbf{(a)} or individual points \textbf{(b-d)}. Black dots are the final node locations and blue lines are entries in the final transition matrix A greater than the initialization value $\frac{1}{N}$.}
    \label{fig:suppA}
\end{figure}

Additionally, we visualize the final node locations and transition entries in $A$ for each experimental dataset in Figure \ref{fig:suppA}. Note that while all timepoints of the data are plotted (grey), only the end result of Bubblewrap is shown (black dots, blue lines). 

\subsection{Degeneration of dynamical systems to random walks}

\begin{figure}
  \centering
  \includegraphics[width=\linewidth]{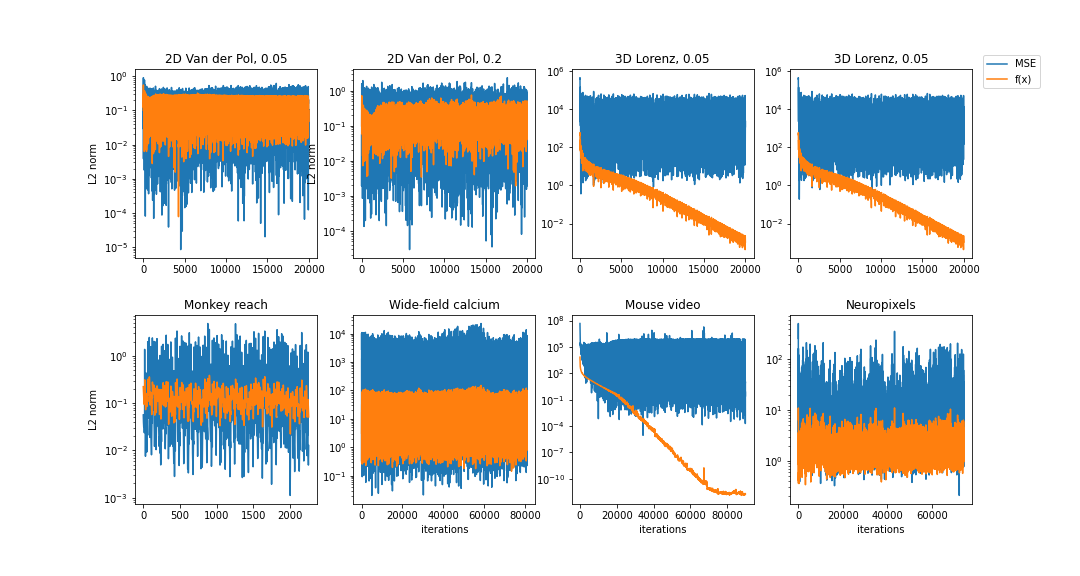}
  \caption{\textbf{ZP2016 degenerates to a random walk for some data sets.} Plots of ZP2016 prediction error (MSE, blue curves) and predicted step size ($f(x)$, orange curves, cf. (\ref{eqn:zp_mean_pred})) for \textbf{a)} 2D Van der Pol (0.05), \textbf{b)} 2D Van der Pol (0.2), \textbf{c)} 3D Lorenz (0.05), \textbf{d)} 3D Lorenz (0.2), \textbf{e)} Monkey reach, \textbf{f)} Wide-field calcium, \textbf{g)} Mouse video, and \textbf{h)} Neuropixels data sets.}
  \label{fig:supfig}
\end{figure}

For the Lorenz and Mouse video datasets, the model of \cite{NIPS2016_b2531e7b} outperformed both Bubblewrap and VJF in log predictive probability. However, as shown in Figure \ref{fig:supfig}, this is because the predicted step size ($f(x)$ in (\ref{eqn:zp_mean_pred})) drops to near 0. That is, the model degenerates to a random walk. Datasets where this occurs are marked with a * in Table \ref{res-table}.

\subsection{Model benchmarking}
\begin{figure}
  \centering
  \includegraphics[width=\linewidth]{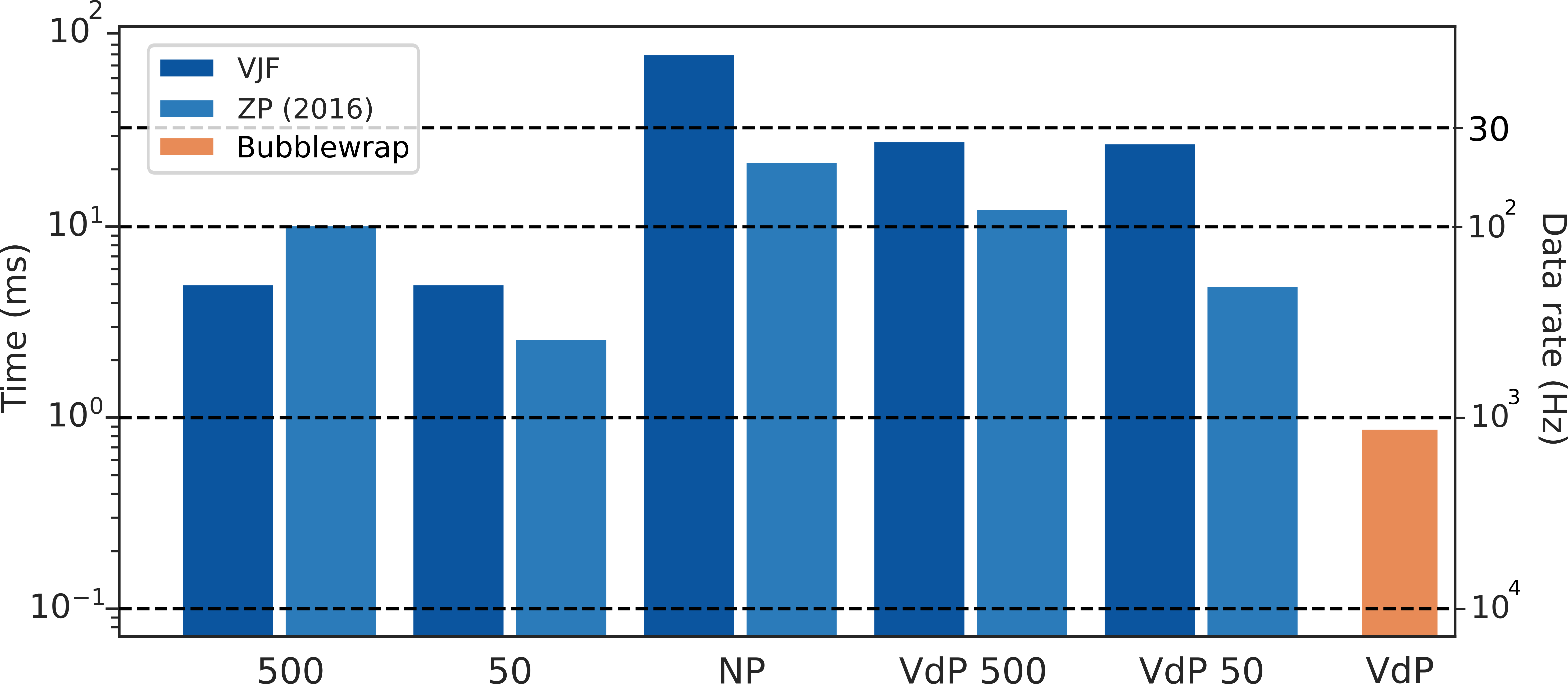}
  \caption{\textbf{Benchmarking of comparison models.} Average runtimes for the time to update the model for 500 (`500') or 50 (`50') basis functions on the 2D Van der Pol (0.05 noise) dataset; for the total time including prediction on the Neuropixels dataset (`NP'),  the 2D Van der Pol dataset with 500 (`VdP 500') or 50 (`Vdp 50') basis functions; and for Bubblewrap on the base case of the 2D Van der Pol dataset using 1k nodes (`VdP').}
  \label{fig:supfigcomp}
\end{figure}

Here we present average runtimes for the comparison models on selected datasets, similar to Figure 4c in the main text. Note that we gave all models the already dimension-reduced data. Figure \ref{fig:supfigcomp} shows only the time needed to update the model for each new datapoint of the 2D Van der Pol (VdP) oscillator dataset for two different numbers of basis functions (labeled `500' and `50'), which are roughly equivalent to the notion of nodes in Bubblewrap. Next, we include the time to generate predictions via sampling, increasing the total computation time, for the Neuropixels dataset (`NP') and the Van der Pol oscillator (0.05 noise), again with 500 or 50 basis functions (`VdP 500' and `VdP 50'). Finally, we plot the average time to fit Bubblewrap (including prediction time) for the base VdP (`VdP') case using 1k nodes. All comparison model cases, using our implementations, showed that models ran at rates slower than 1 kiloHerz.

\subsection{Bubblewrap model evolution}

\begin{figure}
  \centering
  \includegraphics[width=\linewidth]{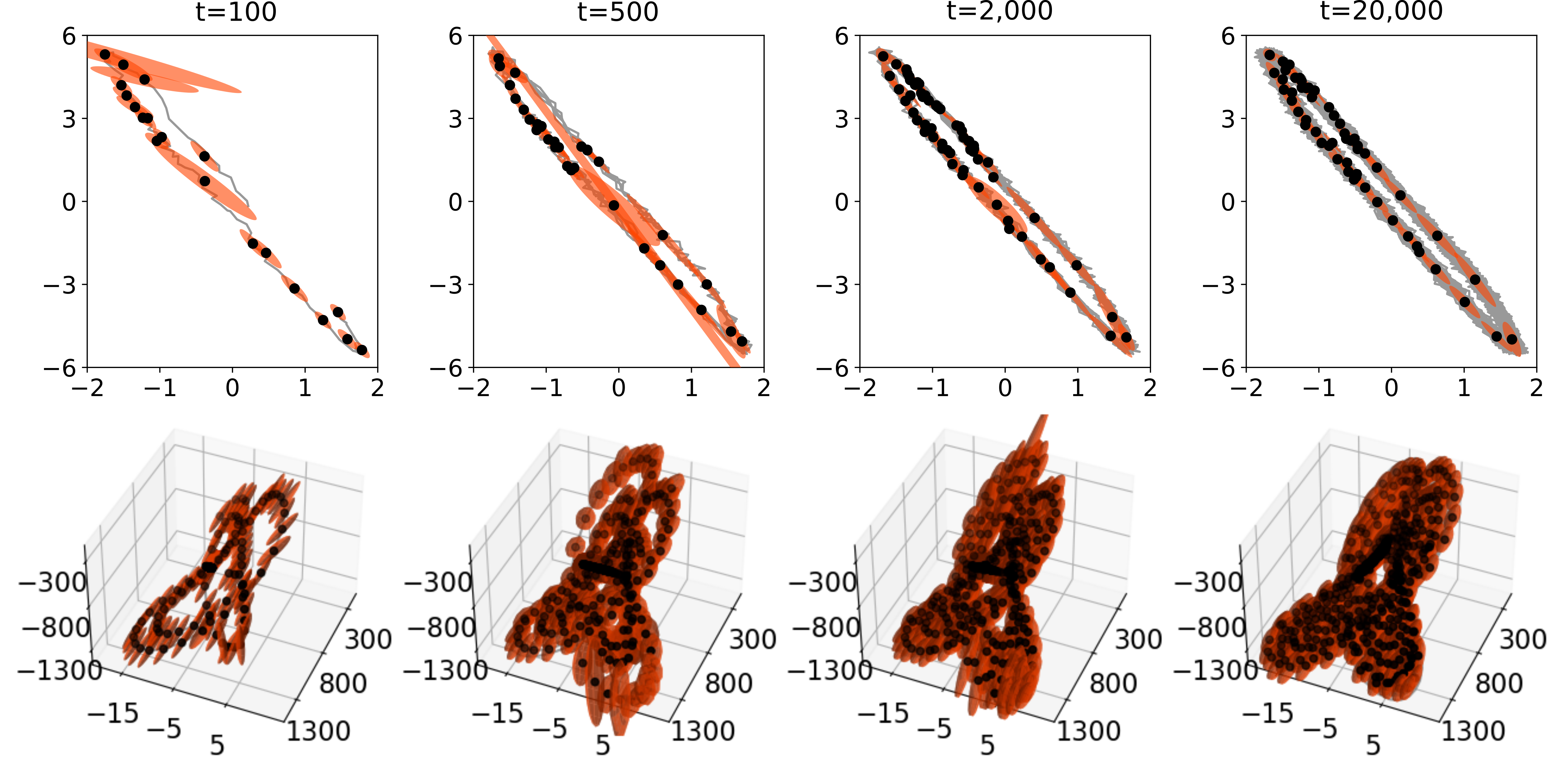}
  \caption{\textbf{Bubblewrap over time for simulated datasets.} Model results after t=100, 500, 2000, and 20000 (entire dataset) data points seen. Top row: 2D Van der Pol; bottom row: 3D Lorenz; both 0.05 noise cases.}
  \label{fig:supptimefig}
\end{figure}

As shown in the above log predictive probability plots, Bubblewrap takes only a small amount of data to successfully fit the underlying neural manifold. Figure \ref{fig:supptimefig} shows our model fit as a function of the number of timepoints seen for the 2D Van der Pol and 3D Lorenz simulated datasets (0.05), with tiling locations settling in after observing a few hundred data points and uncertainty being refined throughout the course of the dataset.

\subsection{Different seeds}
\begin{figure}
  \centering
  \includegraphics[width=\linewidth]{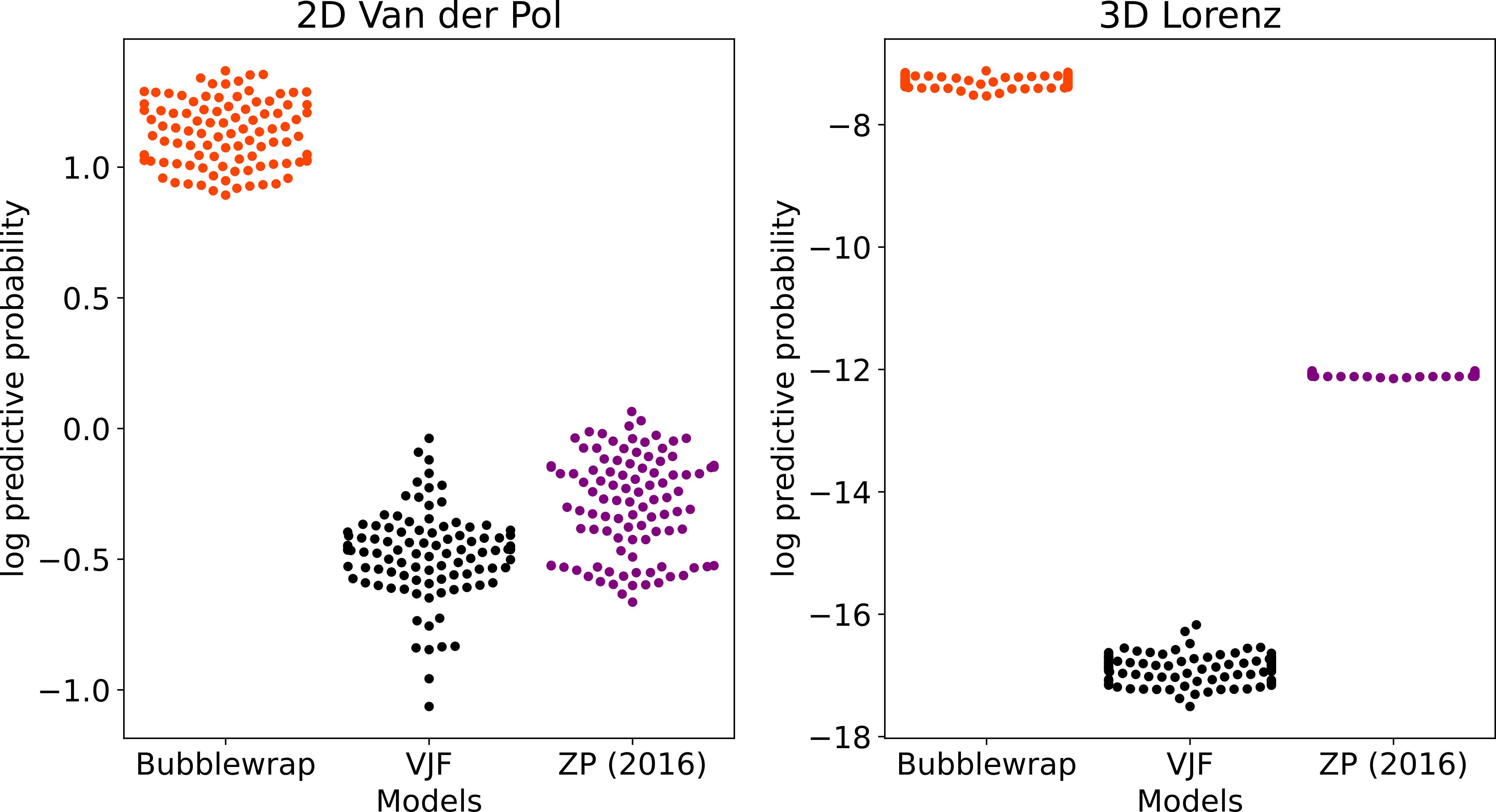}
  \caption{\textbf{Mean log predictive probability across seeds.} Model results for 100 different simulated trajectories of the 2D Van der Pol (0.05 noise) and 3D Lorenz (0.05) datasets.}
  \label{fig:suppseedfig}
\end{figure}

To examine the consistency in the performance of Bubblewrap and the two comparison models, we generated a set of 100 trajectories per dataset using different random seeds to set initial conditions. Figure \ref{fig:suppseedfig} shows the mean log predictive probability for each of these trajectories for the 2D Van der Pol and 3D Lorenz (0.05) datasets.

\subsection{Benchmarking}

\begin{figure}
  \centering
  \includegraphics[width=\linewidth]{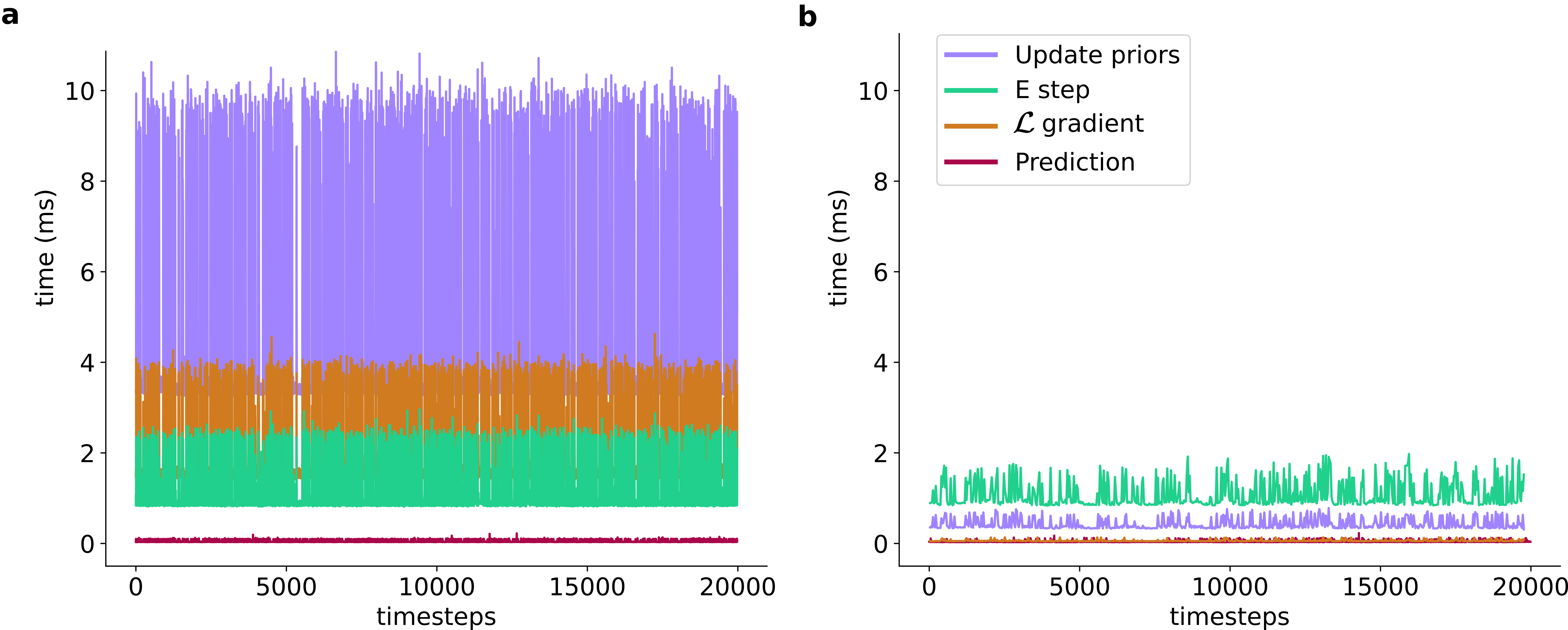}
  \caption{\textbf{Timing for different steps inside Bubblewrap.} \textbf{a)} Raw step times for updating priors, E step, gradient update of $\mathcal{L}$, and prediction for the 2D Van der Pol datatset.
  \textbf{b)} Same as in (a), but in batch mode, where prior updates and the gradient step are performed every 30 points.}
  \label{fig:suppbenchfig}
\end{figure}

Figure \ref{fig:suppbenchfig} displays a typical breakdown of timing for three separate steps within the Bubblewrap algorithm for single step (a) and batch mode (b). For single step updates, the algorithm observes a new data point, updates the data mean and covariance estimates, and uses these to update the node priors (`Update priors'). The algorithm then performs forward filtering and computing sufficient statistics (`E step'). Finally, the M step involves computing and applying gradients of $\mathcal{L}$. (`$\mathcal{L}$ gradient'). In batch mode, prior updates and gradient calculations are performed only once per batch, amortizing the cost of these updates. Finally, we display the time for calculating the model's predicted probability distribution over the tile index at the next time step $p(z_{t+1}|x_{1:t})$ (`Prediction'). This cost is on the order of microseconds.

\subsection{Bubblewrap without heuristics}
\begin{figure}
  \centering
  \includegraphics[width=0.75\linewidth]{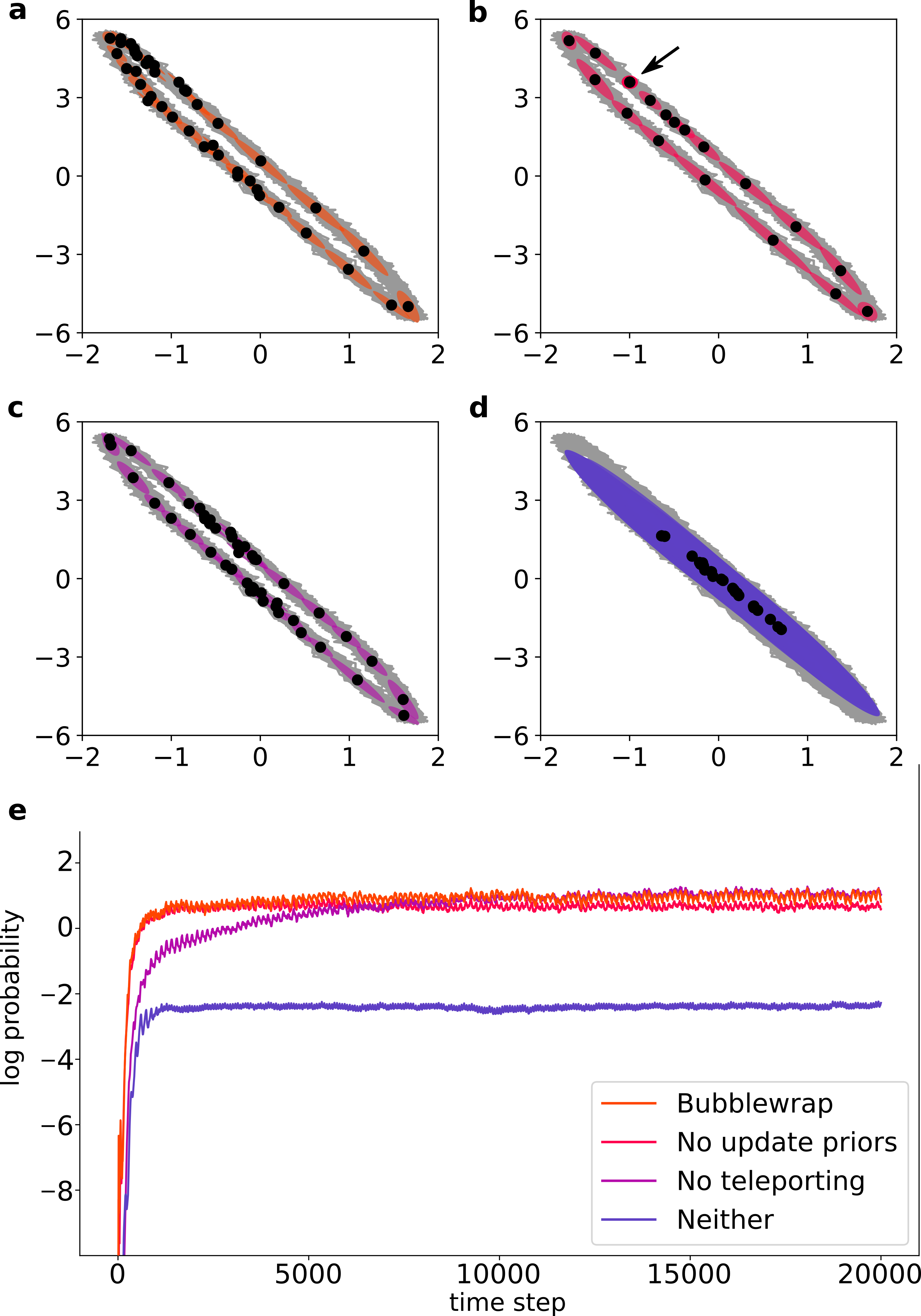}
  \caption{\textbf{Bubblewrap feature comparisons.} Model results for the 2D Van der Pol (0.05 noise) data set shown for different combinations of Bubblewrap heuristics. \textbf{a)} Reproduction of Figure \ref{fig:2d3d}a) with all features enabled. \textbf{b)} Results if priors are not updated with data estimates.
  \textbf{c)} Results if teleporting is not employed.
  \textbf{d)} Results if neither prior updates nor teleporting features are enabled.
  \textbf{e)} Smoothed log predictive probability across all time points for the 4 cases shown above (raw data omitted for clarity).
  }
  \label{fig:suppheurfig}
\end{figure}

As noted in the main text, we employed heuristics in addition to the online EM updates to improve Bubblewrap's performance. Figure \ref{fig:suppheurfig} shows results on the 2D Van der Pol (0.05) dataset with some features removed; 'Bubblewrap' in (a) is plotted for comparison with all features enabled. 

When tiles are initialized, their priors place them at the center of mass of the earliest few data points. With each new data point, we re-estimate the data mean and covariance and use these to update the priors for $\mu_{j}$ and $\Sigma_{j}$. This includes random walk dynamics on $\mu_{0j}$ (cf. (\ref{eqn:mu0_drift})) to break degeneracy among tiles that have seen the same or no data.  If we remove this feature, we are still able to effectively tile the space ('No update priors', Fig. \ref{fig:suppheurfig}b) but a number of redundant, overlapping nodes are left in their initial state (black arrow). For timing considerations, we note that turning off these prior updates provides a speed benefit without drastically diminishing performance in some cases (Fig. \ref{fig:suppheurfig}e).

A second important heuristic we employed was to begin by marking all nodes as available for teleportation and thereby `breadcrumb' the initial set of incoming datapoints. With this teleport feature turned off (`No teleporting', Fig. \ref{fig:suppheurfig}c), again Bubblewrap is able to effectively learn the correct tiling, though at a slower rate during initial learning (Fig. \ref{fig:suppheurfig}e). Additionally, because tile locations must be gradually updated through gradient updates on $\mathcal{L}$ rather than instantaneously updated via teleporting, the entire computation time per new data point is roughly 3 times slower.

Finally, if we employ neither of the above features, we can fall into less-optimal tilings with lower log predictive probability results ('Neither', Fig. \ref{fig:suppheurfig}d,e). Here many nodes have all tried to encompass all of the data simulataneously, with heavily overlapping covariance bubbles.

\end{document}